\documentclass[twoside]{article}

\usepackage[preprint]{aistats2027}

\usepackage[round]{natbib}

\usepackage{algorithm}
\usepackage{algorithmic}
\usepackage{amsfonts}
\usepackage{amsmath}
\usepackage{multirow}
\usepackage{booktabs}
\usepackage{graphicx}
\usepackage{wrapfig}
\usepackage{xcolor}
\usepackage{subcaption}
\usepackage[hyphens]{url}

\newcommand{\scaleboxratio}{0.66}
\newcommand{\norm}[1]{\left \lVert #1 \right \rVert}

\begin{document}

\runningtitle{A Drop-In Study of GNN Layers for DNN Graph Regression}
\twocolumn[

\aistatstitle{How Much Does Message Passing Matter? A Drop-In Study of GNN Layers for Neural Network Graph Regression}

\aistatsauthor{ Keith G. Mills \And Aedan J. DeFrates \And  Joong Ho Kim }

\aistatsaddress{LSU ATHENA Lab, Baton Rouge, LA \\ \{Keith.Mills, Aedan.DeFrates, JoongHo.Kim\}@lsu.edu} 
]

\begin{abstract}
    
Graph Neural Networks (GNNs) are widely used as regressors, for example to predict the accuracy of a neural architecture. %
Yet new message-passing (MP) layers are developed and benchmarked almost exclusively on classification tasks, and regression pipelines typically adopt a single MP layer without ablation. We ask how much the choice of MP layer matters for graph-level regression. Holding the architecture, loss and training recipe of four existing GNN regressors fixed, we substitute ten MP configurations spanning convolutional, isomorphism-based and attention-based designs. %

We evaluate them on eleven datasets of neural-network graphs that range from under ten to over a thousand nodes per graph and from a few hundred to over four hundred thousand samples, measuring rank correlation, prediction error, top-$k$ retrieval, latency and memory. MP choice changes results substantially, and we find that %
the best choice depends on graph size, training-set size and regression objective. Classical layers such as GEN, $k$-GNN and PNA match or exceed attention-based layers on small architecture graphs at lower cost, while GATv2 performs well on datasets with few large graphs. %

\end{abstract}

\section{Introduction}
\label{sec:intro}

Graphs are a popular format for representing irregularly %
structured data, such as neural network designs~\citep{han2023general}. %
The rise of Machine Learning (ML) and the advent of Deep Neural Networks (DNN)~\citep{Krizhevsky09CIFAR} in the 2010s has also paved the way for the development of Graph Neural Networks (GNN), special DNNs which perform ML prediction on graph data~\citep{ju2026graph}. 

Specifically, GNNs are primarily characterized by how they perform message-passing (MP). This %
is the mechanism by which a given node receives information from, and transmits information to, the %
nodes in its neighborhood it shares an edge with. MP mechanisms vary in their scope: One of the oldest mechanisms generalizes the convolution operation to graphs~\citep{welling2016semi}, while newer GNN types implement Transformer attention mechanisms~\citep{vaswani2017attention}, isomorphism tests~\citep{xu2019GIN} including the Weisfeiler-Lehman (WL) test~\citep{morris2019weisfeiler}, etc.~\citep{corso2020principal}. The diversity of MP mechanism design, coupled with the ubiquity of graph data, has allowed GNNs to play an outsized role in advancing not just black-box predictive tasks, but eXplainable AI (XAI)~\citep{ying2019gnnexplainer, luo2020parameterized, lu2024Eig} and Interpretable AI~\citep{pereira2023neural, wan2022Redundant} research as well.

However, advances in ML and DNN research are a numbers~\citep{li2020geometry, howard2019searching, cai2020once, chen2024pixartAlpha} game, where newly proposed methods primarily contribute by usurping older methods on known benchmarks~\citep{mehrotra2021bench}. While GNN MP mechanisms are no exception to this~\citep{chatzianastasis2023GOAT}, an issue lies in that most GNN benchmarks are either node-level or graph-level classification tasks, e.g., binary prediction on whether a molecule is mutagenic~\citep{kazius2005derivation}, can pass through the blood-brain barrier~\citep{wu2018moleculenet}, etc., which only covers one class of problem. 

In contrast, GNNs are fully capable of performing scalar prediction and have done so successfully, specifically in the fields of lightweight prediction for Neural Architecture Search (NAS)~\citep{white2023neural} and other contexts~\citep{heid2021machine, wang2024comprehensive}. Unlike classification, low-cost NAS regression poses its own challenges, including accurately predicting continuous-valued targets and %
preserving the relative rank ordering of samples~\citep{immer2023effective}. Consequently, graph regression commonly relies on a diverse set of evaluation metrics, such as absolute error~\citep{mills2023gennape}, rank correlation~\citep{salameh2024autogo} or relevance-matching~\citep{zhang2021acenas}. Despite being less represented in mainstream GNN literature, these metrics have a higher potential to change as a function of GNN choice compared to their classification counterparts. Moreover, it is often the case that rather than ablating different GNN types, a GNN regression paper will be designed around a specific GNN type~\citep{lu2023pinat, hwang2024flowerformer} or utilize a single GNN type in an off-the-shelf manner~\citep{chen2021contrastive} with little to no ablation even though with modern GNN libraries~\citep{Fey/Lenssen/2019} such a procedure requires but a simple, one-line code change in most scenarios. 
This design defaulting leaves the door open to further performance and efficiency gains as newer GNN types may be able to compensate for overall predictor complexity. 

Thus, the goal of this paper is to study 
to what extent GNN MP layer choice plays a role in the %
performance of GNN regressors. %
Specifically, we perform a systematic, yet straightforward study on multiple existing repositories that propose both simple and complex regression GNNs, controlling only the choice of GNN type in a drop-in manner and then comparing the results. Our detailed contributions are as follows:

First, we consider a breadth of GNN regressors for low-cost NAS ranging from those with simple, stacked GNN layers that take an input graph as-is and create a prediction, to more complex GNN regressors that extract information from %
or alter the adjacency information of graphs. Further, these GNNs vary in terms of %
training losses and primary evaluation metrics. %

Second, we perform a sweep of GNN layers ranging from older convolution-based approaches to isomorphism testing to attention-based methods. Not only do these methods vary in terms of the intention of their MP implementation, they also vary in terms of their computational cost, which provides another avenue for evaluating their effectiveness. %

Third, we consider a variety of different graph datasets, from those that describe the high-level architecture of a DNN to those which capture individual primitive operations. Our benchmarks vary greatly in terms of the number of graphs per dataset, average node and edge statistics and purpose, i.e., from  NAS~\citep{ying2019nasbench101} to DNN compression~\citep{mills2025qua2sedimo}.

Our results show that MP choice substantially changes regression performance but that the best layer depends on the regime. Classical layers such as GEN, $k$-GNN and PNA match or exceed attention-based layers on small architecture graphs at lower latency and memory cost, while multi-head GATv2 performs best on datasets of few, very large graphs.

\section{Background Literature} %
\label{sec:related}

Assume a graph $\mathcal{G}$ with node (vertex) set $\mathcal{V}$ and edge set $\mathcal{E}$. The node set is characterized by a feature matrix $X_\mathcal{V}\in \mathbb{R}^{|\mathcal{V}|\times d_f}$ where $|\mathcal{V}|$ is the number of nodes and $d_f$ is the dataset-dependent number of node features; %
graphs may also possess edge-level features. 

Likewise, %
prediction %
is dataset and task-dependent. Graph-level tasks involve generating a prediction $y_\mathcal{G}'$ over the entire graph $\mathcal{G}$, while node-level prediction tasks~\citep{sen2008collective} make a prediction $y'_{v^{*}}$ on a specific node $v^*\in\mathcal{V}_\mathcal{G}$. In this paper we primarily focus on graph-level tasks.

\subsection{Graph Neural Networks (GNN)}
GNNs facilitate ML-driven prediction on graph data. First, the raw node data $X_\mathcal{V}$ is passed through an initial embedding layer to produce the initial node embeddings $H_\mathcal{V}^0$; each node/row $h_v^0\in\mathbb{R}^{d_0}$ is processed independently here. %

Next, a series of GNN message passing (MP) layers are applied, facilitating the exchange of information from different nodes in accordance to the graph's connectivity. There are many different MP mechanisms %
but they all operate under a similar abstraction. Formally, for the arbitrary $k$-th GNN layer, this is given by 

\begin{equation}
    \centering
    h_v^{k+1}=\texttt{Combine}(h_v^{k},\texttt{Agg}(h_u^k: u \in \mathcal{N}(v))),
    \label{eq:gnn}
\end{equation}
where $\mathcal{N}(v)$ is the local neighborhood of node $v$, enumerated by the index $u$. Importantly, the $\texttt{Combine}$ and $\texttt{Agg}$ (Aggregate) functions are determined by the GNN type. For example, Graph Isomorphism Networks (GIN;~\cite{xu2019GIN}) sum the features of neighboring nodes and apply an MLP $\Theta$ as follows 

\begin{equation}
    \centering
    h_v^{k+1}=\Theta_k((1+\epsilon)h_v^k+\sum_{u\in\mathcal{N}(v)}h_u^k),
    \label{eq:gin}
\end{equation}
where $\epsilon$ is a small constant. 
After a sufficient number $K$ of GNN layers, typically 2 to 8~\citep{you2020design}, we make a prediction. For node-level prediction we may simply extract the last node embedding for a target node $v^*$, $h^K_{v^*}$, and feed it into an MLP to form a prediction  $y'_{v^{*}}$.

Graph-level prediction is a bit more involved as it requires compressing the matrix of node embeddings $H^K_\mathcal{V}$ into a fixed-length vector $h^K_\mathcal{G}$ representing the entire graph. One way to achieve this is by an arithmetic operation like mean as follows 

\begin{equation}
    \centering
    h^K_\mathcal{G}=\dfrac{1}{|\mathcal{V}_\mathcal{G}|}\sum_{v\in\mathcal{V}_\mathcal{G}}h_v^K. 
    \label{eq:mean}
\end{equation}

The graph embedding $h^K_\mathcal{G}$ can then be fed into an MLP to produce a prediction $y_\mathcal{G}'$. Both $y'_{v^{*}}$ and $y_\mathcal{G}'$ can be vectors of classification logits or they can be scalar values for regression, i.e., using the Mean Squared Error (MSE) loss $\mathcal{L}_{MSE}=\mathbb{E}[||y_\mathcal{G}'-y_\mathcal{G}||^2]$. In practice, when comparing different forms of GNN MP mechanisms, classification datasets like Citeseer~\citep{caragea2014citeseer}, Cora, Pubmed~\citep{sen2008collective} and others~\citep{kazius2005derivation, yanardag2015graph} are the de facto benchmarks for evaluating whether one GNN type is superior to another. 

\begin{table*}[t!]
    \centering
    \scalebox{\scaleboxratio}{
    \begin{tabular}{llccccc} \toprule
    \textbf{Repository} & \textbf{Dataset} & \textbf{\#Graphs} & \textbf{Avg. \#Nodes/Edges} & \textbf{$y_\mathcal{G}'$ Distribution \& Range} & \textbf{Metrics} \\ \midrule
    \multirow{5}{*}{\shortstack[l]{FlowerFormer\\\citet{hwang2024flowerformer}}} & NAS-Bench-101 & 14,580 & 7/8.73 & $\mathcal{N}(89.5, 8.0)$; $[9.50, 94.7]$ & \multirow{5}{*}{\shortstack[c]{\textbf{KT}, SRCC,\\MAE, R$^2$}} \\
                                  & NAS-Bench-201 & 15,625 & 8/10 & $\mathcal{N}(87.1, 12.9)$; $[10.0, 94.4]$ & \\
                                  & NAS-Bench-301 & 113,936 & 15/20 & $\mathcal{N}(92.9, 1.4)$; $[72.5, 94.8]$ & \\
                                  & NAS-Bench-ASR & 8,242   & 11/20 & $\mathcal{N}(29.7, 20.6)$; $[19.3, 94.6]$ & \\
                                  & NAS-Bench-Graph & 26,206 & 6/6.10 & $\mathcal{N}(75.4, 9.6)$; $[9.2, 83.1]$ & \\ \midrule
    \multirow{2}{*}{\shortstack[l]{PINAT\\\citet{lu2023pinat}}}        & NAS-Bench-101 & 423,624 & 7/8.73 & $\mathcal{N}(90.2, 5.9
)$; $[9.5, 95.1]$ & \multirow{2}{*}{\shortstack[c]{\textbf{KT}, SRCC,\\MAE, %
    Pr@K}} \\
                                  & NAS-Bench-201 & 15,625 & 8/10 & $\mathcal{N}(83.7, 12.8)$; $[9.71, 91.6]$ & \\ \midrule
    \multirow{5}{*}{\shortstack[l]{AutoBuild\\\citet{mills2024autobuild}}}    & NAS-Bench-101 & 423,624 & 7/8.73 & $\mathcal{N}(90.2, 5.9)$; $[9.5, 95.1]$ & \multirow{5}{*}{\shortstack[c]{\textbf{SRCC,} KT, \\\textbf{MAE,} MAPE}} \\
                                  & NAS-Bench-201 & 15,625 & 8/10 & $\mathcal{N}(87.1, 12.9)$; $[10.0, 94.7]$ & \\
                                  & OFA-MBv3 & 3,000 & 14.97/13.96 & $\mathcal{N}(76.9, 0.81)$; $[74.2, 78.7]$ & \\
                                  & OFA-PN & 3,000 & 16.10/15.10 & $\mathcal{N}(75.4, 0.86)$; $[72.3, 77.4]$ & \\
                                  & MBV3-PN-ONNX-IR & 6,000 & 225.18/249.69 & $\mathcal{N}(76.2, 1.11)$; $[72.3, 78.7]$ & \\ \midrule
    \multirow{3}{*}{\shortstack[l]{Qua$^2$SeDiMo\\\citet{mills2025qua2sedimo}}} 
    & PixArt-$\alpha$ & 374 & 568/818 & $\mathcal{N}(318, 123)$; $[92, 508]$ & \multirow{3}{*}{\shortstack[c]{\textbf{SRCC, NDCG,}\\ KT, MAE}} \\
                                  & HunYuan-DiT & 340 & 844/1172 & $\mathcal{N}(292, 75)$; $[164, 480]$ & \\
                                  & Stable Diffusion XL &   447 & 1,316/1,701 & $\mathcal{N}(348, 137)$; $[102, 705]$ \\
                                  \bottomrule

    \end{tabular}
    }
    \caption{Summary of %
    graph regression datasets utilized. %
    We consider four NAS predictor repositories that utilize GNNs for regression and their supported datasets. For each dataset we report the number of graphs, average number of nodes/edges across the dataset, as well as the distribution and range of target labels. Finally, we also enumerate the regression metrics each repository considers for predictors. Targets refer to the performance of a DNN design the graph reflects; we refer the reader to the original paper for details. 
    \textbf{Bolded} metrics are primary to evaluation we consider them when conducting statistical significance tests. %
    }
    \label{tab:datasets}
    \vspace{-2mm}
\end{table*}

\subsubsection{Regression with GNNs.} 
Scalar estimation and ranking poses different challenges compared to classification. Some regressors prioritize the exactness of predictions through error metrics like Mean Absolute Error (MAE), etc., while others focus on rank correlation metrics like Kendall's Tau (KT) or Spearman Rank Correlation Coefficient (SRCC). Others further still may recast prediction in the context of Information Retrieval (IR) metrics like Normalized Discounted Cumulative Gain (NDCG)~\citep{zhang2021acenas}. 
This paper aims to explore the efficacy of different GNN MP mechanisms in these scenarios and verify whether or not gains made in the classification domain hold for regressor GNNs. 

\subsection{Design Space Benchmarks for GNNs}
Prior work has aimed to perform broad-scale ablation 
of GNN components. %
The most noteworthy work is likely \citet{you2020design}, who consider 315k GNN designs over 32 different classification tasks. Another more recent work is \citet{luo2024classic} who focus on simple GNN ablations for node classification and who find that simpler, classical designs can still provide SOTA performance compared to newer graph transformers. In contrast, %
we focus specifically on GNN MP choice across a variety of pre-existing NAS predictor %
repositories in terms of their original overall GNN architecture and training recipes. To the best of our knowledge, no such %
study has been conducted.

\section{Datasets and Predictors}
\label{sec:method}

Table~\ref{tab:datasets} enumerates the repositories and %
datasets we consider. %
These datasets stem primarily from Neural Architecture Search (NAS) efficient performance predictor~\citep{white2021powerful} development. Unless otherwise stated, we %
utilize the graph datasets made available with the existing open-source repository and re-run predictor experiments as-is using the predefined hyperparameters. In the supplementary, we elaborate on these parameters as well as consider parameter-matched scenarios. %

\subsection{FlowerFormer}

FlowerFormer~\citep{hwang2024flowerformer} serves as our first regression environment and represents the most architecturally complex predictor in this study. 
It captures information flows within the Directed Acyclic Graph (DAG) representation of a DNN architecture by modeling both the forward-pass and backward-pass information to update node features. This predictor is coupled with an underlying DAGFormer~\citep{luo2023transformers} GNN MP mechanism, which is a Graph Transformer~\citep{yun2019graph} designed for DAG data. FlowerFormer has $K=2$ GNN layers. 

FlowerFormer covers several popular datasets such as NAS-Bench-101 (NB-101;~\cite{ying2019nasbench101}), NAS-Bench-201 (NB-201;~\cite{dong2020nasbench201}), NAS-Bench-301 (NB-301;~\cite{zela2022surrogate}), NAS-Bench-ASR (NB-ASR;~\cite{mehrotra2021bench}) and NAS-Bench-Graph (NB-Graph;~\cite{qin2022bench}). %
FlowerFormer trains on a pairwise ranking loss

\begin{equation}
    \centering
    \label{eq:ff_loss}
    \mathcal{L}_{pw} = \sum_{(i, j):y_{\mathcal{G}_i} > y_{\mathcal{G}_j}}max(0, \texttt{margin}-(y_{\mathcal{G}_i}'-y_{\mathcal{G}_j}')),
\end{equation}
where $\texttt{margin}\in(0, 1]$ is a small scalar; this loss aims to maximize ranking evaluation metrics like Kendall's Tau (KT) primarily. %
Finally, the FlowerFormer evaluation is staged with different proportions of the dataset utilized to train the predictor. %

\subsection{PINAT}
PINAT~\citep{lu2023pinat} serves as a second Graph Transformer-based environment emphasizing 
permutation invariance between different data graphs. To facilitate this, the initial node embedding layer features a transform on the graph laplacian, followed by a Permutation Invariance Module (PIM) applied to %
the initial node embedding, %
in order to extract information and differentiate different graphs. The PINAT GNN contains $K=3$ custom `GATSet' MP layers; we refer the reader to the original paper for details. 

PINAT %
supports running experiments on NB-101 and NB-201. %
PINAT trains using the Mean Squared Error (MSE) loss augmented by a sequential ranking loss~\citep{yi2023nar} as 

\begin{equation}
    \centering
    \label{eq:pinat_loss}
    \mathcal{L}_{P} = \mathcal{L}_{\mathrm{MSE}} + \lambda \sum_{i=1}^{n} \left| \left(y_{I(i)}'-y_i'\right) - \left(y_{I(i)}-y_i\right) \right|,
\end{equation}
where $I(i)$ is a randomly sampled architecture and $\lambda$ is a weighting coefficient. The primary evaluation metric to consider for PINAT is KT. %

Like FlowerFormer, PINAT evaluation is performed in stages where different proportions of the dataset are utilized to train the predictor. A key research question this paper aims to answer is to what extent the complex graph encodings performed by FlowerFormer and PINAT require complex Graph Transformers compared to older, but less resource-intense GNN MP mechanisms. 

\subsection{AutoBuild}
This is our third GNN predictor environment. Structurally, AutoBuild is 
simpler than that of FlowerFormer and PINAT: It consists of an initial node embedding layer, followed by several GNN MP layers, graph embedding aggregation of Eq.~\ref{eq:mean}, and then prediction; there is no complex transform computed on the laplacian and the adjacency information is kept as-is. The %
GNN has $K=4$ layers.

However, the complexity of AutoBuild comes from the loss function it utilizes, which incorporates the differentiable SRCC loss $\mathcal{L}_{rank}$ from \citet{blondel2020fast} in addition to an MSE loss as 

\begin{equation}
    \centering
    \label{eq:ab_loss}
    \mathcal{L}_{AB} = \mathcal{L}_{MSE} + \dfrac{1}{K+1}\sum_{k=0}^{K}\mathcal{L}_{rank}(y_\mathcal{G}, \norm{h_\mathcal{G}^k}_1),
\end{equation}
to create a predictor that correlates intermediate graph embedding norms with the target at each hop-level; %
we refer the reader to the original paper for further details. As a result, AutoBuild primarily focuses on ranking SRCC as well as MAE. %

Unlike other works, AutoBuild focuses on performance prediction for MobileNet variants, such as Once-for-All (OFA) MobileNetv3 (MBv3)~\citep{cai2020once} and ProxylessNAS (PN) MobileNetv2 (MBv2)~\citep{cai2018proxylessnas}, %
though we %
extend this %
repository to facilitate prediction on NB-101, NB-201 and NB-301. %

AutoBuild is where we start to transition from datasets that contain a lot of small graphs, akin to molecules or chemical substances, to datasets that have fewer graphs but where the average graph contains over 100 nodes. Specifically, per Tab.~\ref{tab:datasets}, the graphs for `OFA-MBv3' and `OFA-PN' %
are simple sequence graphs that reflect the high-level neural architecture design decisions, and are akin to the graphs for NAS-Benchmarks. In contrast, `MBV3-PN-ONNX-IR' (abbrev. `ONNX-IR' for ONNX-Intermediate Representation) graphs reflect the actual, primitive operation-level representation of the neural network as it would exist in ONNX~\citep{bai2019Onnx} %
and %
require %
more nodes and details to represent the same architectures.

\subsection{Qua$^2$SeDiMo}

This regressor environment applies NAS principles and concepts from AutoBuild to facilitate Diffusion Model (DM)~\citep{podell2023sdxl} post-training quantization~\citep{li2023model}. %
For each dataset, the graph structure itself, including number of nodes and connectivity, is the same, however, the node features change across each dataset entry describing how it is to be quantized. Like `MBV3-PN-ONNX-IR', the graphs are large as they reflect the entire structure of the neural network, not simply high-level design decisions. Qua$^2$SeDiMo uses Eq.~\ref{eq:ab_loss} as a loss but extends $\mathcal{L}_{rank}$ to include LambdaRank~\citep{burges2010ranknet} for NDCG in addition to the differentiable SRCC. 

Moreover, %
the target label is not accuracy but the Fr\'echet Inception Distance (FID;~\citep{heusel2017gans}) of a quantized DM variant. %
There are %
fewer graphs in each %
dataset compared to others, making these benchmarks the extreme case for predicting a regression target from a large number of node features while having an extremely limited number of training data and labels to work with, which is a good stress test. %

\begin{figure*}[t!]
    \centering
    \includegraphics[width=0.97\linewidth]{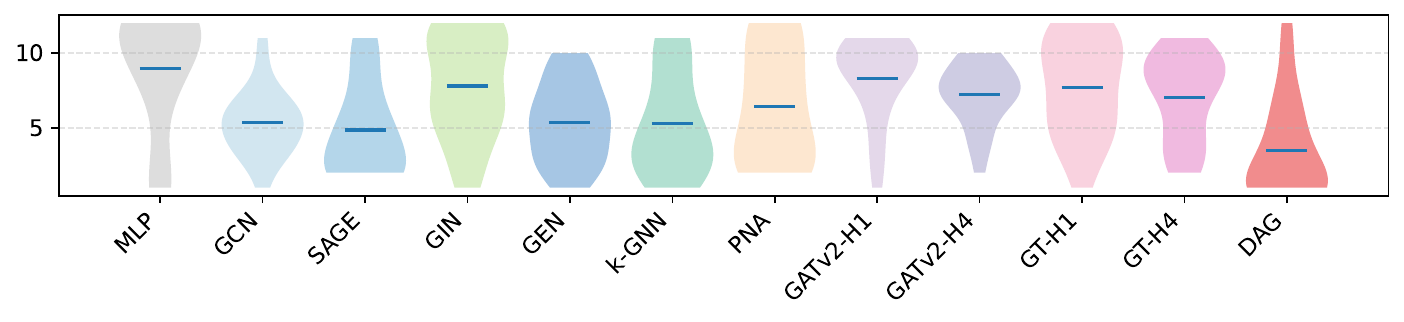}
    \vspace{-5mm}
    \caption{Violin plot of the GNN ranking distributions for %
    FlowerFormer KT results %
    including mean (blue line). Y-Axis indicates rank between 1 and 12. Lower rank is better. Best viewed in color.}
    \label{fig:violin_ff}
\end{figure*}
\begin{table*}[t!]
    \centering
    \scalebox{\scaleboxratio}{
    \begin{tabular}{llccccccccccccc} \toprule
    \textbf{Dataset} & \textbf{Train \%} & \textbf{MLP} & \textbf{GCN} & \textbf{SAGE} & \textbf{GIN} &  \textbf{GEN} & \textbf{$k$-GNN} & \textbf{PNA} & \textbf{GATv2-H1} & \textbf{GATv2-H4} & \textbf{GT-H1} & \textbf{GT-H4} & \textbf{DAG} \\ \midrule
\multirow{4}{*}{NB-101} & KT-1\% & $0.35_{0.04}$ & $0.72_{0.02}$ & $0.71_{0.03}$ & $0.68_{0.02}$ & $0.72_{0.02}$ & $\mathit{0.74}_{0.02}$ & $\underline{0.75}_{0.03}$ & $0.72_{0.04}$ & $0.72_{0.02}$ & $\mathit{0.74}_{0.02}$ & $\mathit{0.74}_{0.01}$ & $\mathbf{0.78}_{0.02}$ \\
& KT-5\% & $0.40_{0.01}$ & $\mathit{0.84}_{0.00}$  & $\underline{0.85}_{0.01}$ & $\mathit{0.84}_{0.01}$ & $\mathit{0.84}_{0.00}$ & $\underline{0.85}_{0.00}$ & $\underline{0.85}_{0.00}$ & $\mathit{0.84}_{0.01}$ & $\mathit{0.84}_{0.00}$ & $\mathit{0.84}_{0.00}$ & $\underline{0.85}_{0.00}$ & $\mathbf{0.86}_{0.01}$ \\
& KT-10\% & $0.40_{0.02}$ & $\underline{0.87}_{0.00}$ & $\underline{0.87}_{0.00}$ & $\underline{0.87}_{0.00}$ & $\underline{0.87}_{0.00}$ & $\underline{0.87}_{0.00}$ & $\underline{0.87}_{0.00}$ &  $\mathit{0.86}_{0.00}$ & $\mathit{0.86}_{0.00}$ &$\mathit{0.86}_{0.00}$ & $\underline{0.87}_{0.00}$ & $\mathbf{0.88}_{0.00}$ \\
& KT-50\% & $\mathit{0.42}_{0.00}$ & $\mathbf{0.90}_{0.00}$ & $\mathbf{0.90}_{0.00}$ & $\underline{0.89}_{0.00}$ & $\underline{0.89}_{0.00}$ & $\mathbf{0.90}_{0.00}$ & $\underline{0.89}_{0.00}$ &  $\underline{0.89}_{0.00}$ & $\underline{0.89}_{0.00}$ & $\underline{0.89}_{0.00}$ & $\underline{0.89}_{0.00}$ & $\underline{0.89}_{0.00}$ \\ \midrule

\multirow{4}{*}{NB-201} &KT-1\% & $\mathit{0.71}_{0.04}$ & $\mathit{0.71}_{0.02}$ & $\underline{0.72}_{0.02}$ & $0.68_{0.03}$ & $\mathit{0.71}_{0.01}$ & $\underline{0.72}_{0.02}$ & $\mathit{0.71}_{0.02}$ & $0.69_{0.03}$ & $\mathit{0.71}_{0.03}$ & $0.67_{0.02}$ & $0.70_{0.01}$ & $\mathbf{0.77}_{0.01}$ \\
& KT-5\% & $\mathit{0.85}_{0.01}$ & $\underline{0.86}_{0.01}$ & $\underline{0.86}_{0.01}$ & $0.84_{0.01}$ & $\underline{0.86}_{0.01}$ & $\underline{0.86}_{0.00}$ & $\underline{0.86}_{0.00}$ & $\mathit{0.85}_{0.01}$ & $0.86_{0.00}$ & $0.83_{0.01}$ & $0.84_{0.00}$ & $\mathbf{0.89}_{0.01}$ \\
& KT-10\% & $0.88_{0.00}$ & $\underline{0.90}_{0.00}$ & $\mathbf{0.91}_{0.00}$ &$\mathit{0.89}_{0.01}$ & $\underline{0.90}_{0.00}$ & $\underline{0.90}_{0.00}$ & $\underline{0.90}_{0.00}$ & $\mathit{0.89}_{0.00}$ & $\underline{0.90}_{0.00}$ & $0.88_{0.01}$ & $0.88_{0.00}$ & $\mathbf{0.91}_{0.00}$ \\
& KT-50\% & $\underline{0.89}_{0.00}$ & $\mathbf{0.93}_{0.00}$ & $\mathbf{0.93}_{0.00}$ & $\mathbf{0.93}_{0.00}$ & $\mathbf{0.93}_{0.00}$ & $\mathbf{0.93}_{0.00}$ & $\mathbf{0.93}_{0.00}$ & $\mathbf{0.93}_{0.00}$ & $\mathbf{0.93}_{0.00}$ & $\mathbf{0.93}_{0.00}$ & $\mathbf{0.93}_{0.00}$ & $\mathbf{0.93}_{0.00}$ \\  \midrule

\multirow{4}{*}{NB-301} &KT-1\% & $0.50_{0.14}$ & $0.60_{0.03}$ & $\underline{0.62}_{0.01}$ & $0.59_{0.01}$& $0.49_{0.21}$ & $\mathbf{0.63}_{0.02}$ & -- & $\underline{0.62}_{0.01}$ & $0.54_{0.12}$ & $0.32_{0.17}$ & $0.32_{0.17}$ & $\mathit{0.61}_{0.02}$ \\
& KT-5\% & $0.16_{0.12}$ & $0.22_{0.09}$ & $0.14_{0.10}$ & $0.22_{0.08}$ & $\mathbf{0.37}_{0.08}$ & $\underline{0.35}_{0.08}$ & -- & $0.17_{0.11}$ & $0.20_{0.09}$ & $0.22_{0.05}$ & $0.19_{0.04}$ & $\mathit{0.26}_{0.02}$ \\
& KT-10\% & $0.06_{0.14}$ & $0.26_{0.10}$ & $0.25_{0.10}$ & $\mathbf{0.37}_{0.07}$ & $\underline{0.34}_{0.10}$ & $\underline{0.34}_{0.07}$ & -- & $0.02_{0.06}$ & $0.17_{0.13}$ & $0.27_{0.08}$ & $\mathit{0.30}_{0.05}$ & $0.28_{0.06}$ \\
& KT-50\% & $0.22_{0.04}$ & $0.36_{0.08}$ & $0.36_{0.09}$ & $\underline{0.39}_{0.09}$ & $\textit{0.37}_{0.06}$ & $\mathbf{0.59}_{0.05}$ & -- & $0.02_{0.17}$ & $0.21_{0.09}$ & $0.28_{0.04}$ & $0.31_{0.06}$ & $0.28_{0.08}$ \\ \midrule

\multirow{4}{*}{NB-ASR} & KT-1\% & $\underline{0.32}_{0.04}$ & $0.30_{0.07}$ & $\mathit{0.31}_{0.03}$ & $\underline{0.32}_{0.02}$ &$0.29_{0.05}$ & $0.28_{0.04}$ & $0.29_{0.06}$ & $\mathbf{0.33}_{0.04}$ & $\mathbf{0.33}_{0.04}$ & $0.29_{0.03}$ & $0.29_{0.03}$ & $0.24_{0.07}$ \\
& KT-5\% & $\mathbf{0.46}_{0.02}$ & $\mathit{0.42}_{0.01}$ & $\underline{0.43}_{0.01}$ & $0.39_{0.01}$ & $\mathit{0.42}_{0.02}$ & $0.40_{0.02}$ & $0.41_{0.02}$ & $\mathit{0.42}_{0.01}$ & $\mathit{0.42}_{0.01}$ & $0.40_{0.02}$ & $0.40_{0.02}$ & $0.41_{0.03}$ \\
& KT-10\% & $\mathbf{0.49}_{0.01}$ & $0.46_{0.01}$ & $\mathit{0.47}_{0.02}$ & $0.44_{0.01}$ & $\mathit{0.47}_{0.02}$ & $0.45_{0.01}$ & $\mathit{0.47}_{0.02}$ & $0.46_{0.01}$ & $0.46_{0.02}$ & $\mathit{0.47}_{0.02}$ & $0.46_{0.02}$ & $\underline{0.48}_{0.01}$ \\
& KT-50\% & $\mathbf{0.59}_{0.00}$ & $0.52_{0.01}$ & $\mathit{0.54}_{0.02}$ & $0.51_{0.01}$ & $0.52_{0.01}$ & $0.52_{0.01}$ & $\underline{0.58}_{0.01}$ & $0.53_{0.03}$ & $0.52_{0.00}$ & $\underline{0.58}_{0.01}$ & $\underline{0.58}_{0.01}$ & $0.53_{0.01}$ \\ \midrule

\multirow{4}{*}{NB-Graph} & KT-1\% & $0.41_{0.02}$ & $\mathit{0.46}_{0.02}$ & $0.42_{0.03}$ & $0.45_{0.01}$ & $\mathit{0.46}_{0.01}$ & $0.45_{0.01}$ & $\underline{0.48}_{0.02}$ & $0.42_{0.02}$ & $0.44_{0.03}$ & $0.42_{0.01}$ & $0.44_{0.05}$ & $\mathbf{0.51}_{0.01}$ \\
& KT-5\% & $0.55_{0.01}$ & $\mathit{0.61}_{0.01}$ & $0.60_{0.01}$ & $0.60_{0.01}$ &$\mathit{0.61}_{0.01}$ & $0.60_{0.01}$ & $\underline{0.62}_{0.01}$ &  $0.59_{0.01}$ & $\mathit{0.61}_{0.01}$ & $0.59_{0.03}$ & $0.59_{0.02}$ & $\mathbf{0.65}_{0.01}$\\
& KT-10\% & $0.59_{0.01}$ & $\mathit{0.66}_{0.01}$ & $\mathit{0.66}_{0.01}$ & $0.65_{0.01}$ & $\mathit{0.66}_{0.01}$ & $\mathit{0.66}_{0.00}$ & $\underline{0.67}_{0.01}$ & $\mathit{0.66}_{0.00}$ & $\mathit{0.66}_{0.01}$ & $\mathit{0.66}_{0.01}$ & $\mathit{0.66}_{0.01}$ & $\mathbf{0.69}_{0.00}$ \\
& KT-50\% & $0.65_{0.00}$ & $\mathbf{0.74}_{0.00}$ & $\mathbf{0.74}_{0.00}$ & $\underline{0.73}_{0.00}$ & $\mathbf{0.74}_{0.00}$ & $\mathbf{0.74}_{0.00}$ & $\mathit{0.66}_{0.16}$ & $\mathbf{0.74}_{0.00}$ & $\mathbf{0.74}_{0.00}$ & $\mathbf{0.74}_{0.00}$ & $\mathbf{0.74}_{0.00}$ & $\underline{0.73}_{0.00}$ \\ \bottomrule
    \end{tabular}
    }
    \caption{FlowerFormer %
    KT results. Rows correspond to %
    training-set fractions %
    utilized for training. Horizontal lines demarcate different datasets. %
    Results averaged across 5 random seeds with standard deviation as a subscript. `--' indicates OOM error and defaults to the worst rank for Fig.~\ref{fig:violin_ff}.}
    \label{tab:ff_results}
    \vspace{-3mm}
\end{table*}

\section{Experimental Results}
\label{sec:results}

We enumerate the GNN layers we consider. %
Then, we %
provide %
results and analysis of our study across different repositories and then provide a hardware cost analysis of the different GNN types. 
For all tabular performance results, we report the \textbf{best}, \underline{second best} and \textit{third best} GNN MP mechanisms with \textbf{bold}, \underline{underlined} or %
\textit{italicized} text, respectively. Due to space constraints, the main text reports primary metrics only. The supplementary material contains secondary metrics (SRCC, MAE, R$^2$, MAPE and Pr@10), AutoBuild hop-level SRCC, the full significance-test outputs, parameter-matched experiments, and additional hardware measurements. 

\subsection{GNN MP Mechanisms}
We consider a non-MP MLP baseline that simply applies a Linear, BatchNorm and ReLU operation sequence to each node feature independently. In terms of traditional GNN MP mechanisms, we consider drop-in MP mechanisms that are plug-and-play from PyTorch-Geometric (PyG;~\cite{Fey/Lenssen/2019}): 
Graph Convolution (GCN;~\cite{welling2016semi}), %
GraphSAGE~\citep{hamilton2017inductive}, Graph Isomorphism (GIN;~\cite{xu2019GIN}), GEN (DeeperGCN;~\cite{li2020deepergcn}), $k$-GNN~\citep{morris2019weisfeiler} and Principal Neighborhood Aggregation (PNA;~\cite{corso2020principal}).

We also consider %
attention-based MP mechanisms such as Graph Attention v2 %
(GATv2;~\cite{brody2022how}) and PyG's %
Graph Transformer (GT;~\cite{shi2020masked}, which applies multi-head attention over each node's local neighborhood without positional encodings. For each of these, we consider whether the attention mechanism has 1 or 4 heads. Finally, for all but one of these GNNs, using them as a drop-in replacement is a simple code change; the exception is PNA which requires a degree histogram computed once from the training data.

\subsection{Results on Complex Regressors} %
\label{sec:ff_and_pinat}

We turn our attention towards FlowerFormer and PINAT as the most architecturally complex NAS GNN regressors. Specifically, we focus on the primary evaluation metric of these predictors, Kendall's Tau (KT), and use the training-set sizes of the original paper.

\subsubsection{Results on FlowerFormer.}

FlowerFormer consists of 5 datasets multiplied by 4 percentage-based training dataset splits: 1\%, 5\%, 10\% and 50\%, giving rise to 20 different trials. 
Figure~\ref{fig:violin_ff} provides a summary violin plot of the results of these trials. The overall best GNN is the default DAGFormer which achieves the lowest mean rank of 3.5 compared to other GNNs. Surprisingly though, in this drop-in experimental setting, the other attention-based GNNs lag behind DAGFormer and rank closer to the MLP baseline than to GCN, GEN, GraphSAGE or $k$-GNN. %
Meanwhile, GNNs based on convolutions, such as GCN and GEN, as well as those rooted in graph theory like GraphSAGE and $k$-GNN achieve more competitive performance. 

Table~\ref{tab:ff_results} further details the exact numerical results that comprise Fig.~\ref{fig:violin_ff}. %
While the superior performance of DAGFormer is primarily concentrated in NB-101, NB-201 and NB-Graph, it falls short on NB-301,
which is dominated by GIN, GEN and $k$-GNN, as well as NB-ASR where at the lowest training data size, GATv2 achieves the best performance. Interestingly, KT on NB-301 does not improve monotonically with training data for most layers, including DAGFormer, suggesting instability under the repository's training recipe; we report it as-is. Also, %
surprisingly, the simple MLP baseline outperforms all GNN MP mechanisms when more training data is permitted on NB-ASR. Furthermore, when DAGFormer excels, it does so at lower training set sizes. As we can see for NB-101, NB-201 and NB-Graph, %
the performance gap is decreased and DAGFormer may be surpassed by the time we reach 50\% training data set size.

Further, we turn our attention towards the cases where DAGFormer or the MLP achieves the best performance and ask what is the 2nd or 3rd best GNN? For NB-101, %
PNA achieves a consistent 2nd place %
across data splits, while GraphSAGE, $k$-GNN and Graph Transformer (GT) also do well with 5\% data or more. %
Convolution-based or isomorphism GNNs, i.e., GCN, GEN and GIN require more data to perform as well on NB-101, but are more competitive on NB-201. %
PNA does well on NB-Graph %
once the data exceeds 5\%, but causes a CUDA OOM error on NB-301. %
Moreover, unlike other GNN MP mechanisms, PNA requires additional graph information and is not a simple drop-in module. Besides PNA, GCN and GEN achieve 3rd best performance on NB-Graph and then outperform PNA and DAGFormer when 50\% data is available.

\subsubsection{Results on PINAT.}
\begin{figure*}[t!]
    \centering
    \includegraphics[width=0.97\linewidth]{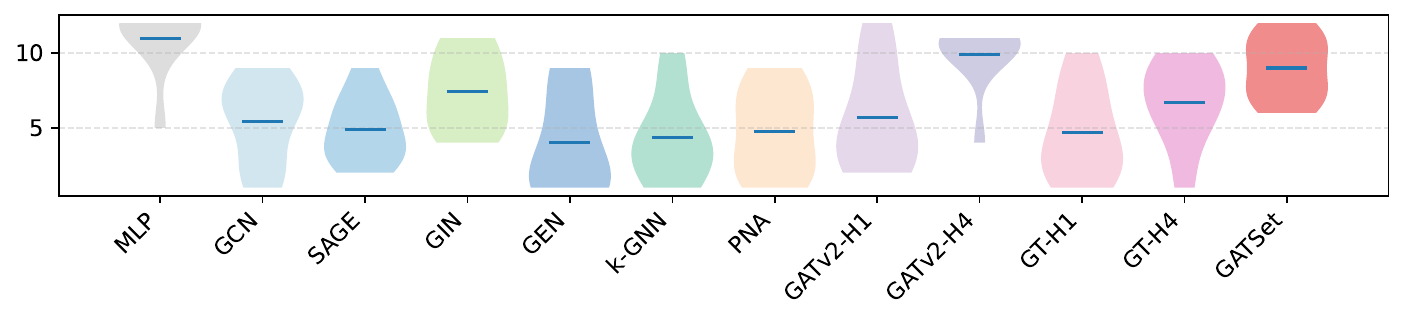}
    \vspace{-5mm}
    \caption{Violin plot of the GNN ranking distributions for %
    PINAT KT results %
    including mean (blue line). Y-Axis indicates rank between 1 and 12. Lower rank is better. Best viewed in color.}
    \label{fig:violin_pinat}
\end{figure*}

Next, Figure~\ref{fig:violin_pinat} repeats the violin plot for PINAT, where we have nine trials: 4 for NB-101 and 5 for NB-201, per the original paper. Here, the custom GATSet not only replaces DAGFormer but, in contrast to it, achieves very low performance compared to \textit{all} but one of the off-the-shelf drop-in replacements we consider in this paper. Specifically, the top-5 MP mechanisms for PINAT are GEN, $k$-GNN, PNA, GT-H1 and GraphSAGE, roughly in that order.

\begin{table*}[t!]
    \centering
    \scalebox{\scaleboxratio}{
    \begin{tabular}{llcccccccccccc} \toprule
    \textbf{Dataset} & \textbf{Train \#} & \textbf{MLP} & \textbf{GCN} & \textbf{SAGE} & \textbf{GIN} &  \textbf{GEN} & \textbf{$k$-GNN} & \textbf{PNA} & \textbf{GATv2-H1} & \textbf{GATv2-H4} & \textbf{GT-H1} & \textbf{GT-H4} & \textbf{GATSet} \\ \midrule
\multirow{4}{*}{NB-101} & KT-100 & $0.61_{0.01}$ & $\mathit{0.67}_{0.01}$ & $\mathit{0.67}_{0.03}$ & $0.64_{0.01}$ & $\mathit{0.67}_{0.01}$ & $\mathbf{0.70}_{0.01}$ & $\underline{0.69}_{0.03}$ & $\mathit{0.67}_{0.02}$ & $0.64_{0.01}$ & $\underline{0.69}_{0.02}$ & $\mathit{0.67}_{0.03}$ & $0.65_{0.02}$ \\
& KT-172 & $0.66_{0.01}$ & $\mathit{0.72}_{0.00}$ & $\underline{0.73}_{0.01}$ & $0.71_{0.01}$ & $\mathit{0.72}_{0.01}$ & $0.71_{0.01}$ & $\mathit{0.72}_{0.01}$ & $\mathit{0.72}_{0.01}$ & $\underline{0.73}_{0.01}$ & $\mathbf{0.74}_{0.00}$ & $\mathbf{0.74}_{0.01}$ & $0.71_{0.01}$ \\
& KT-424 & $0.70_{0.01}$ & $0.75_{0.00}$ & $\mathit{0.76}_{0.00}$ & $\mathit{0.76}_{0.01}$ & $\underline{0.77}_{0.00}$ & $\mathit{0.76}_{0.00}$ & $\mathbf{0.78}_{0.01}$ & $0.75_{0.00}$ & $0.74_{0.01}$ & $\mathit{0.76}_{0.01}$ & $\mathit{0.76}_{0.01}$ & $\mathit{0.76}_{0.01}$ \\
& KT-4236 & $0.81_{0.00}$ & $\underline{0.85}_{0.00}$ & $\mathit{0.84}_{0.00}$ & $\underline{0.85}_{0.00}$ & $\underline{0.85}_{0.00}$ & $\mathbf{0.86}_{0.00}$ & $\underline{0.85}_{0.00}$ & $\mathit{0.84}_{0.00}$ & $\mathit{0.84}_{0.00}$ & $\mathit{0.84}_{0.00}$ & $\mathit{0.84}_{0.00}$ & $\underline{0.85}_{0.00}$ \\ \midrule 

\multirow{5}{*}{NB-201} & KT-78 & $0.09_{0.32}$ & $\mathbf{0.29}_{0.27}$ & $\underline{0.24}_{0.19}$ & $0.03_{0.38}$ & $-0.04_{0.24}$ & $\mathit{0.21}_{0.18}$ & $0.07_{0.32}$ & $-0.26_{0.26}$ &  $-0.15_{0.23}$ & $-0.04_{0.26}$ & $0.13_{0.18}$ & $-0.03_{0.28}$ \\
& KT-156 & $0.51_{0.03}$ & $\mathbf{0.55}_{0.04}$ & $\mathit{0.53}_{0.02}$ & $\mathbf{0.55}_{0.04}$ & $\mathbf{0.55}_{0.05}$ & $\mathit{0.53}_{0.03}$ & $\mathit{0.53}_{0.04}$ & $\mathbf{0.55}_{0.02}$ &  $0.52_{0.04}$ & $\mathbf{0.55}_{0.03}$ & $\mathit{0.53}_{0.03}$ & $0.52_{0.04}$ \\
& KT-469 & $0.62_{0.01}$ & $0.66_{0.02}$ & $\mathit{0.67}_{0.01}$ & $0.66_{0.01}$ & $\mathbf{0.69}_{0.01}$ & $\underline{0.68}_{0.01}$ & $\mathit{0.67}_{0.02}$ & $\mathit{0.67}_{0.02}$ & $0.64_{0.02}$ & $\mathit{0.67}_{0.02}$ & $0.66_{0.02}$ & $0.63_{0.02}$ \\
& KT-781 & $0.69_{0.01}$ & $\mathit{0.72}_{0.01}$ & $\underline{0.73}_{0.01}$ & $\mathit{0.72}_{0.01}$ & $\mathbf{0.74}_{0.01}$ & $\underline{0.73}_{0.00}$ & $\underline{0.73}_{0.01}$ & $\underline{0.73}_{0.01}$ & $0.71_{0.01}$ & $\underline{0.73}_{0.01}$ & $\mathit{0.72}_{0.00}$ & $0.69_{0.02}$ \\
& KT-1563 & $0.75_{0.02}$ & $\mathit{0.76}_{0.01}$ & $\underline{0.77}_{0.01}$ & $\mathit{0.76}_{0.01}$ & $\mathbf{0.79}_{0.01}$ & $\mathit{0.76}_{0.01}$ & $\mathit{0.76}_{0.01}$ & $\underline{0.77}_{0.01}$ & $0.75_{0.01}$ & $\underline{0.77}_{0.01}$ & $\mathit{0.76}_{0.01}$ & $0.74_{0.01}$ \\ 
\bottomrule

    \end{tabular}
    }
    \caption{PINAT Kendall's Tau (KT) results. Same setup as Tab.~\ref{tab:ff_results} except PINAT uses sample counts instead of training-set proportions. %
    `GATSet' is custom to PINAT.}
    \label{tab:pinat_results}
    \vspace{-2mm}
\end{table*}

\begin{figure*}[t!]
    \centering
    \includegraphics[width=0.90\linewidth]{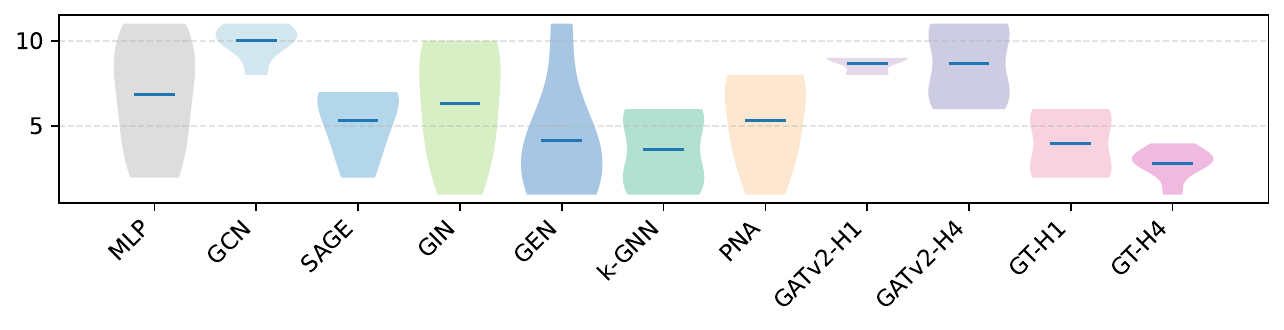}
    \vspace{-5mm}
    \caption{Violin plot of the GNN ranking distributions for %
    AutoBuild MAE results %
    including mean (blue line). Y-Axis indicates rank between 1 and 11. Lower rank is better. Best viewed in color.}
    \label{fig:violin_ab_mae}
\end{figure*}

\begin{table*}[t!]
    \centering
    \scalebox{\scaleboxratio}{
    \begin{tabular}{lcccccccccccc} \toprule
    \textbf{Dataset} & \textbf{MLP} & \textbf{GCN} & \textbf{SAGE} & \textbf{GIN} &  \textbf{GEN} & \textbf{$k$-GNN} & \textbf{PNA} & \textbf{GATv2-H1} & \textbf{GATv2-H4} & \textbf{GT-H1} & \textbf{GT-H4} \\ \midrule
NB-101 & $0.34_{0.01}$ & $0.22_{0.02}$ & $\underline{0.16}_{0.00}$ & $\mathit{0.18}_{0.01}$ & $\mathbf{0.15}_{0.00}$ & $\mathbf{0.15}_{0.00}$ & $\mathbf{0.15}_{0.01}$ &  $0.20_{0.00}$ & $\underline{0.16}_{0.01}$ & $\underline{0.16}_{0.01}$ & $\mathbf{0.15}_{0.01}$ \\ 
NB-201 & $0.21_{0.03}$ & $0.40_{0.25}$ & $0.15_{0.05}$ & $\mathit{0.13}_{0.02}$ & $\mathbf{0.07}_{0.01}$ & $0.15_{0.05}$ & $\underline{0.09}_{0.03}$ &  $0.21_{0.08}$ & $0.49_{0.23}$ & $\underline{0.09}_{0.02}$ & $\underline{0.09}_{0.02}$ \\ 
NB-301 & $\mathit{0.27}_{0.01}$ & $0.43_{0.09}$ & $0.29_{0.05}$ & $\mathit{0.27}_{0.04}$ & $\mathbf{0.25}_{0.01}$ & $\underline{0.26}_{0.02}$ & $0.32_{0.09}$ & $0.33_{0.04}$ & $0.34_{0.10}$ & $0.28_{0.01}$ & $\mathit{0.27}_{0.02}$ \\
OFA-MBv3 & $\underline{0.30}_{0.04}$ & $0.34_{0.03}$ & $0.32_{0.05}$ & $0.39_{0.08}$ & $0.39_{0.08}$ & $0.32_{0.05}$ & $0.33_{0.05}$ & $0.35_{0.04}$ & $0.32_{0.05}$ & $\mathit{0.31}_{0.04}$ & $\mathbf{0.29}_{0.05}$ \\

OFA-PN & $\mathit{0.15}_{0.02}$ & $0.19_{0.01}$ & $\mathbf{0.12}_{0.02}$ & $0.19_{0.02}$ & $\underline{0.14}_{0.01}$ & $\mathbf{0.12}_{0.02}$ & $0.16_{0.05}$ & $0.17_{0.02}$ & $0.16_{0.02}$ & $\underline{0.14}_{0.02}$ & $\underline{0.14}_{0.02}$ \\

ONNX-IR & $0.19_{0.04}$ & $0.27_{0.09}$ & $0.18_{0.06}$ & $\mathbf{0.12}_{0.02}$ & $\mathit{0.15}_{0.03}$ & $\mathit{0.15}_{0.03}$ & $\underline{0.13}_{0.03}$ & $0.18_{0.03}$ & $0.57_{0.62}$ & $\mathbf{0.12}_{0.01}$ & $\underline{0.13}_{0.03}$ \\ \bottomrule

    \end{tabular}
    }
    \caption{AutoBuild MAE results. Lower is better. Rows correspond to different datasets. 
    Results averaged across 5 random seeds.}
    \label{tab:autobuild_mae}
    \vspace{-4mm}
\end{table*}

Continuing, Table~\ref{tab:pinat_results} provides specific numerical results across the nine PINAT trials. %
Results on NB-101 %
are chaotic: $k$-GNN achieves the best performance at the lowest and highest data splits, while PNA and GT take the intermediate data splits. The story is clearer on NB-201. %
When given a low amount of training data, the results are dominated by seed variance. However, a traditional GCN is very data efficient and achieves respectable KT. Increase the data set size more and GATv2/GT %
become more competitive. Give the predictor a lot of data to train on, and GEN, which is a GCN designed for deeper GNNs, will do very well. 

Both FlowerFormer and PINAT %
couple complex graph transforms %
with advanced MP mechanisms like %
DAGFormer %
to facilitate prediction. We examine the performance of these predictors when the GNN MP mechanism is swapped out for an alternative. %
In the case of FlowerFormer, %
DAGFormer is %
powerful and excels in low-data scenarios, %
the case is not %
the same for the custom GATSet of PINAT. %

\subsection{Results on AutoBuild}

Figure~\ref{fig:violin_ab_mae} and Table~\ref{tab:autobuild_mae} report the violin plots and numerical AutoBuild MAE results, respectively. 
Immediately, three types of GNNs stand out: GEN, %
$k$-GNN and %
Graph Transformers (GT), as achieving low average rank, i.e., high average performance. %
GEN dominates performance on the trio of NAS-Benchmarks while GT-H1 and GT-H4 perform well on OFA-MBv3, PN and ONNX-IR which (ONNX-IR especially) contain larger graph data samples, but are smaller in terms of the total number of graphs. Additional SRCC results are provided in the supplementary. %

\begin{table*}[t!]
    \centering
    \scalebox{\scaleboxratio}{
    \begin{tabular}{llcccccccccccc} \toprule
    \textbf{DM Dataset} & \textbf{Metric} & \textbf{MLP} & \textbf{GCN} & \textbf{SAGE} & \textbf{GIN} &  \textbf{GEN} & \textbf{$k$-GNN} & \textbf{PNA} & \textbf{GATv2-H1} & \textbf{GATv2-H4} & \textbf{GT-H1} & \textbf{GT-H4} \\ \midrule

PixArt-$\alpha$ & \multirow{3}{*}{SRCC} & $0.03_{0.03}$ & $-0.36_{0.03}$ & $-0.18_{0.05}$ & $-0.46_{0.03}$ & $\mathbf{0.47}_{0.04}$ & $-0.49_{0.02}$ & $\mathit{0.20}_{0.04}$ & $-0.47_{0.05}$ & $\underline{0.34}_{0.02}$ & $0.13_{0.01}$ & $-0.12_{0.02}$ \\

Hunyuan & & $0.69_{0.04}$ & $0.67_{0.03}$ & $0.60_{0.09}$ & $\mathit{0.70}_{0.02}$ & $\underline{0.72}_{0.04}$ & $0.68_{0.03}$ & $0.59_{0.06}$ & $0.66_{0.06}$ & $\mathbf{0.74}_{0.02}$ & $0.64_{0.08}$ & $0.69_{0.03}$ \\

SDXL & & NaN & $-0.08_{0.03}$ & $0.04_{0.03}$ & $\mathit{0.40}_{0.04}$ & $0.31_{0.01}$ & $-0.43_{0.04}$ & $0.23_{0.04}$ & $\underline{0.60}_{0.05}$ & $\mathbf{0.71}_{0.02}$ & $-0.09_{0.03}$ & $0.23_{0.03}$ \\

\midrule

PixArt-$\alpha$ & \multirow{3}{*}{NDCG} & $0.59_{0.05}$ & $0.62_{0.09}$ & $0.41_{0.05}$ & $0.67_{0.07}$ & $\mathit{0.69}_{0.07}$ & $0.59_{0.10}$& $0.45_{0.06}$ & $\underline{0.70}_{0.11}$ & $\mathbf{0.76}_{0.15}$ & $0.50_{0.06}$ & $0.68_{0.10}$ \\

Hunyuan & & $0.80_{0.07}$ & $0.79_{0.07}$ & $0.77_{0.09}$ & $\mathit{0.83}_{0.08}$ & $\mathbf{0.87}_{0.04}$ & $\mathit{0.83}_{0.09}$ & $0.78_{0.07}$ & $0.78_{0.07}$ & $\underline{0.84}_{0.05}$ & $0.81_{0.07}$ & $\mathit{0.83}_{0.07}$ \\

SDXL & & NaN & $0.38_{0.06}$ & $0.40_{0.05}$ & $\mathit{0.66}_{0.05}$ & $0.61_{0.03}$ & $0.21_{0.03}$ & $0.60_{0.05}$ & $\underline{0.78}_{0.02}$ & $\mathbf{0.84}_{0.06}$ & $0.41_{0.04}$ & $0.53_{0.04}$ \\

\bottomrule
    \end{tabular}
    }
    \caption{Qua$^2$SeDiMo SRCC and NDCG Results. We average the SRCC and NDCG performance across hop-levels $k\in\{1, 2, 3, 4\}$. Higher is better. %
    Horizontal lines demarcate different datasets. %
    Results averaged across 5 random seeds. NaN results defaulted the worst rank where applicable.}
    \label{tab:qua2sedimo_srcc_ndcg}
\end{table*}

\subsection{Results on Qua$^2$SeDiMo}

We stress test our GNNs by considering the case of Qua$^2$SeDiMo, where the GNN regressor %
is working with a small dataset of extremely large and complex graphs and prediction is more difficult. %
Recall the characteristics of Qua$^2$SeDiMo datasets from Tab.~\ref{tab:datasets}: Each contains a small number of very large graphs. Thus, 
compressing such a high amount of graph data down to single scalar values is a daunting task, especially with such a small number of scalars available.

We provide SRCC and NDCG results in Table~\ref{tab:qua2sedimo_srcc_ndcg}. GATv2-H4 stands out as %
consistently %
strong in this situation for both SRCC and NDCG. Further, GATv2-H1 achieves the 2nd best performance on SDXL for SRCC and NDCG, and PixArt-$\alpha$ for NDCG. The other GNN that does well on these datasets is GEN on PixArt-$\alpha$ and Hunyuan, but not SDXL. The advantage is specific to GATv2 rather than to attention in general: GT, which also uses multi-head attention, performs poorly on these datasets. Because this regime is represented by only three datasets, we view it as a %
case where GATv2 is preferable. %

\subsection{Aggregate Comparison and Robustness}
To test whether these rank differences hold across benchmarks rather than within individual trials, we collapse each regressor-dataset pair into a single averaged rank per GNN over its training splits and primary metrics, yielding 16 blocks. %
A Friedman test over these blocks rejects the null hypothesis of equal ranks ($\chi^2 = 31.41$, $p = 5.0 \times 10^{-4}$). GEN attains the best mean rank (3.12), followed by GT-H4 (4.94), PNA (5.03) and $k$-GNN (5.28). However, a post-hoc Nemenyi test finds GEN significantly better only than the MLP, GCN and GATv2-H1 ($p < 0.05$); its advantage over GT, PNA, $k$-GNN and GraphSAGE is not statistically distinguishable. Weighting each dataset equally yields the same conclusion, and under parameter matching on three AutoBuild datasets, MAE rankings are moderately preserved, although on ONNX-IR resizing any layer away from the repository's tuned width degrades it (Supp.~Sec.~\ref{sec:stats} and Tab.~\ref{tab:autobuild_compute_match_mae}).

\subsection{Hardware Analysis}

Finally, we provide %
insight on the hardware trade-offs of %
different GNN types. %
Specifically, we contrast benchmark performance with inference latency and peak memory on an RTX Pro 5000 card. Figure~\ref{fig:hw} illustrates the results for FlowerFormer on NAS-Bench-101 with additional results %
in the supplementary. The results provide an additional facet of analysis: 
While DAGFormer achieves the highest performance, this comes at the cost of significantly increased inference latency and noticeably larger memory footprint than other GNNs. In contrast, $k$-GNN achieves a solid 2nd place in many ways: 2nd highest performance, over $8$x reduction in latency and smaller memory footprint. %

\begin{figure}
    \centering
    \includegraphics[width=\linewidth]{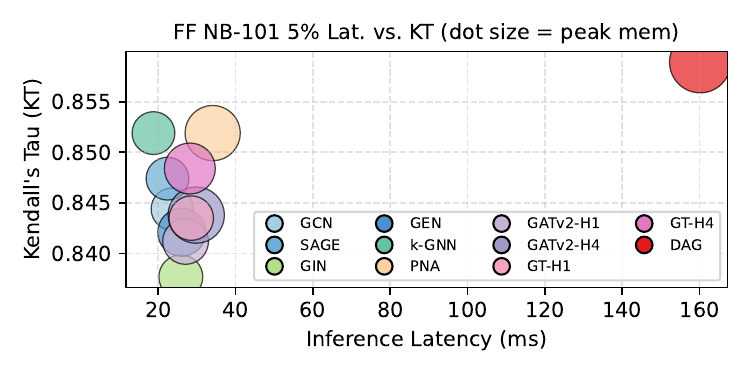}
    \vspace{-4mm}
    \caption{Latency vs. KT plot for FlowerFormer NB-101 5\% %
    training data. Icon size indicates peak memory usage. Best viewed in color.}
    \label{fig:hw}
\end{figure}

\section{Conclusions \& Limitations }
\label{sec:conclusion}

We study the impact of MP layer choice %
for graph-level regression for NAS DNN performance predictors by substituting ten MP configurations into four existing GNN regressors while holding their architectures, losses and training recipes fixed. MP choice changes results substantially, but no single layer dominates. GEN attains the best mean rank across regressor-dataset pairs, yet is statistically distinguishable only from the weakest layers, while $k$-GNN, PNA and GraphSAGE remain competitive at lower latency and memory cost. GATv2 with four heads is strongest on the few-sample, thousand-node diffusion-model graphs of Qua$^2$SeDiMo. A repository's default layer is not always the best choice: DAGFormer leads FlowerFormer at a substantial hardware cost, whereas PINAT's GATSet is outperformed by most drop-in alternatives.

Our conclusions carry two limitations: 1) Retaining each repository's hyperparameters measures MP layers as drop-in replacements rather than at their individually tuned optimum; %
2) All datasets describe DNN architectures, so we do not claim the rankings transfer to molecular or other graph regression domains. %

\newpage
\bibliography{aistats2027}

\clearpage
\appendix
\thispagestyle{empty}

\onecolumn

\appendix
\section{Supplementary Materials}
\label{sec:supp}

\subsection{Regressor Hyperparameters}
We now enumerate the hyperparameters utilized for each GNN regressor repository. Unless otherwise noted, we do not change these hyperparameters, and instead simply change the form of GNN MP mechanism utilized. 

\subsubsection{FlowerFormer.} The regressor GNN utilizes a ReLU activation with `Add' aggregation on the node embeddings to form the graph embedding. The AdamW optimizer is utilized with gradient clipping, however, the exact value of other predictor hyperparameters, e.g., embedding dimension $d$ and the training hyperparameters varies with each dataset as follows:

\begin{enumerate}
    \item \textbf{NB-101:} Trains for 300 epochs with a batch size of 128, initial learning rate of $8e^{-4}$ and weight decay of $3e^{-5}$. Embedding dimension is 64.
    \item \textbf{NB-201:} Trains for 600 epochs with a batch size of 128, initial learning rate of $8e^{-4}$ and weight decay of $6e^{-5}$. Embedding dimension is 64.
    \item \textbf{NB-301:} Trains for 500 epochs with a batch size of 64, initial learning rate of $3e^{-3}$ and weight decay of $2e^{-5}$. Embedding dimension is 256.
    \item \textbf{NB-ASR:} Trains for 1000 epochs with a batch size of 32, initial learning rate of $9e^{-4}$ and weight decay of $9e^{-4}$. Embedding dimension is 128.
    \item \textbf{NB-Graph:} Trains for 200 epochs with a batch size of 1024, initial learning rate of $7e^{-3}$ and weight decay of $4.5e^{-10}$. Embedding dimension is 256.
\end{enumerate}

\subsubsection{PINAT.} The hyperparameters for this repository are the same for both NB-101 and NB-201. The regressor trains for 300 epochs with a batch size of 10, an initial learning rate of $1e^{-4}$ and a weight decay of $1e^{-3}$ using the Adam optimizer. The GNN embedding dimension is $d=256$. Like FlowerFormer, the train/test split is dynamic and all data not in the training split is in the test split.

\subsubsection{AutoBuild.} Each predictor trains for 200 epochs with a batch size of 128, initial learning rate of $1e^{-4}$ with a weight decay of $1e^{-5}$ using the AdamW optimizer. The GNN embedding dimension is $d=32$. Data is split 90\%/10\% into train and test sets for each dataset. We set the target equation to be $y=acc$.

\subsubsection{Qua$^2$SeDiMo.} Each predictor trains for 10000 epochs with a batch size of 128, initial learning rate of $1e^{-3}$ with a weight decay of $1e^{-6}$ using the AdamW optimizer. The GNN embedding dimension is $d=64$. Data is split 80\%/20\% into train and test sets for each dataset. We set the target equation to be $y=-\texttt{FID}$ and utilize Qua$^2$SeDiMo's hybrid loss that combines SRCC and LambdaRank.

\subsection{AutoBuild SRCC Results}

\begin{figure*}[t!]
    \centering
    \includegraphics[width=0.90\linewidth]{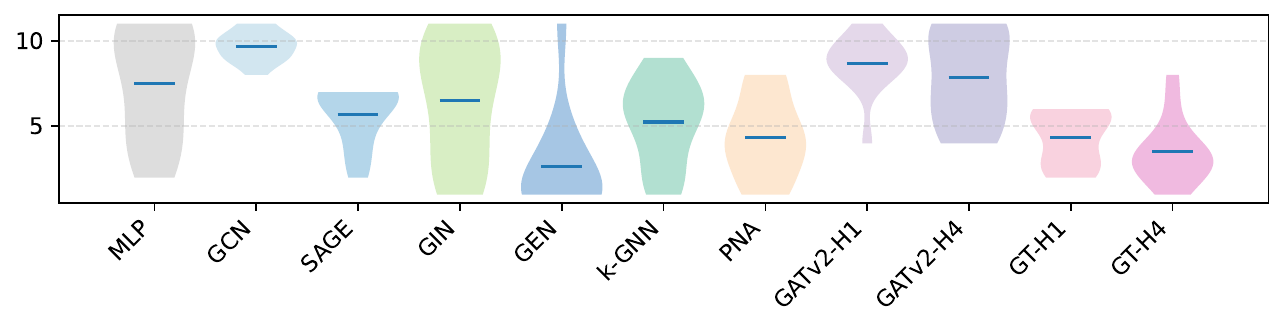}
    \vspace{-5mm}
    \caption{Violin plot of the GNN ranking distributions for %
    AutoBuild SRCC results %
    including mean (blue line). Y-Axis indicates rank between 1 and 11. Lower rank is better. Best viewed in color.}
    \label{fig:violin_ab_srcc}
\end{figure*}

\begin{table*}[t!]
    \centering
    \scalebox{\scaleboxratio}{
    \begin{tabular}{lcccccccccccc} \toprule
    \textbf{Dataset} & \textbf{MLP} & \textbf{GCN} & \textbf{SAGE} & \textbf{GIN} &  \textbf{GEN} & \textbf{$k$-GNN} & \textbf{PNA} & \textbf{GATv2-H1} & \textbf{GATv2-H4} & \textbf{GT-H1} & \textbf{GT-H4} \\ \midrule

NB-101 & $0.62_{0.01}$ & $0.73_{0.02}$ & $0.82_{0.03}$ & $0.81_{0.01}$ & $\mathbf{0.89}_{0.01}$ & $\mathit{0.84}_{0.01}$ & $\mathit{0.84}_{0.04}$ & $0.75_{0.03}$ & $0.81_{0.04}$ & $\mathit{0.84}_{0.01}$ & $\underline{0.86}_{0.01}$ \\

NB-201 & $0.72_{0.03}$ & $0.64_{0.08}$ & $0.71_{0.07}$ & $0.65_{0.09}$ & $\mathbf{0.82}_{0.05}$ & $0.70_{0.06}$& $0.74_{0.09}$ & $0.58_{0.15}$ & $0.52_{0.15}$ & $\mathit{0.76}_{0.09}$ & $\underline{0.78}_{0.08}$ \\

NB-301 & $\mathit{0.78}_{0.01}$ & $0.74_{0.02}$ & $0.77_{0.03}$ & $\underline{0.80}_{0.01}$ & $\mathbf{0.82}_{0.01}$ & $0.76_{0.02}$ & $\mathit{0.78}_{0.03}$ & $0.74_{0.04}$ & $\mathit{0.78}_{0.04}$ & $0.77_{0.02}$ & $0.76_{0.03}$ \\

OFA-MBv3 & $0.83_{0.06}$ & $0.83_{0.04}$ & $0.84_{0.06}$ & $0.80_{0.07}$ & $\mathbf{0.87}_{0.04}$ & $0.84_{0.06}$& $\mathbf{0.87}_{0.04}$ & $0.82_{0.04}$ & $\mathit{0.85}_{0.04}$ & $\underline{0.86}_{0.04}$ & $\mathit{0.85}_{0.06}$ \\

OFA-PN & $0.93_{0.04}$ & $0.94_{0.01}$ & $\underline{0.97}_{0.01}$ & $0.94_{0.03}$ & $\mathbf{0.98}_{0.01}$ & $0.95_{0.01}$ & $\underline{0.97}_{0.01}$ & $0.94_{0.01}$ & $\mathit{0.96}_{0.01}$ & $\mathit{0.96}_{0.01}$ & $0.95_{0.01}$ \\

ONNX-IR & $0.88_{0.05}$ & $0.91_{0.02}$ & $0.92_{0.02}$ & $\underline{0.95}_{0.02}$ & $\mathbf{0.98}_{0.00}$ & $0.91_{0.01}$ & $\mathit{0.94}_{0.02}$ & $\mathit{0.94}_{0.02}$ & $0.91_{0.05}$ & $0.93_{0.02}$ & $\mathit{0.94}_{0.02}$ \\

\bottomrule
    \end{tabular}
    }
    \caption{AutoBuild SRCC Results. We average the SRCC performance across hop-levels $k\in\{1, 2, 3, 4\}$. Higher is better. Rows correspond to different datasets.  %
    Results averaged across 5 random seeds.}
    \label{tab:autobuild_srcc}
    \vspace{-4mm}
\end{table*}

AutoBuild utilizes a differentiable ranking loss to correlate graph embedding norms to target labels; 
we refer interested readers to \citet{mills2024autobuild} for details and motivation. This loss is evaluated at different hop $k$ levels. Figure~\ref{fig:violin_ab_srcc} and Table~\ref{tab:autobuild_srcc} provides the results when we average the results for hop-levels $k\in\{1, 2, 3, 4\}$ in terms of a violin plot and numerical values, respectively. %
GEN dominates performance on this test across all benchmark datasets. PNA ties GEN on OFA-MBv3 but that is the only first-placed tie made. Second and third place results are scattered, mostly between GIN, PNA and GT, and to a less extent, GraphSAGE, $k$-GNN and GATv2. Ironically, while GEN excels at this task, the vanilla GCN falls short and is outperformed by the MLP baseline in several respects. %

\begin{wrapfigure}{rt}{0.5\textwidth}
    \centering
    \includegraphics[width=\linewidth]{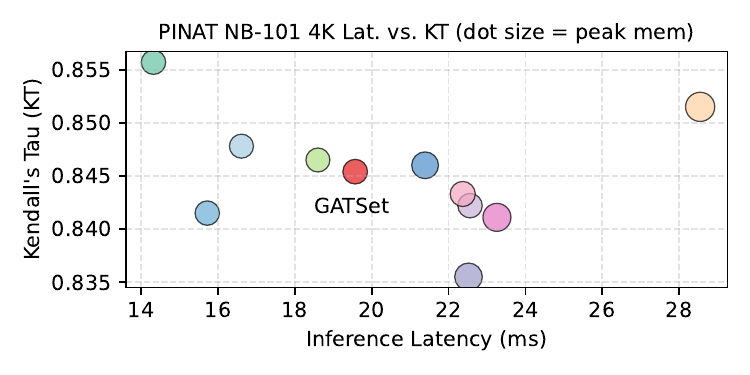}
    \centering
    \includegraphics[width=\linewidth]{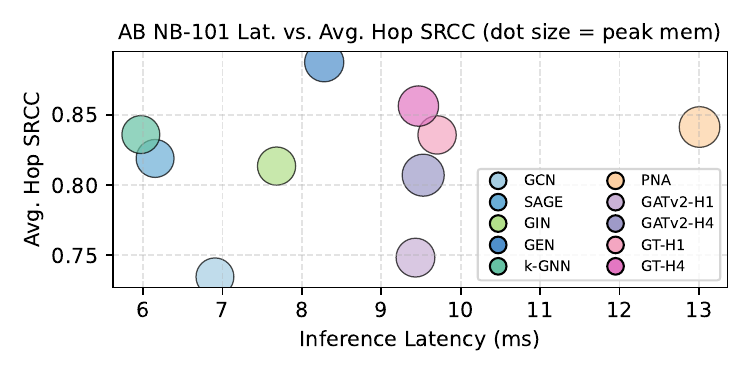}
    \caption{Latency vs. KT plots for PINAT NB-101 4k (upper) and latency vs. Hop SRCC for AutoBuild NB-101 (lower).     
    Icon size indicates peak memory usage. Best viewed in color.}
    \label{fig:supp_hw}
    \vspace{-12mm}
\end{wrapfigure}
Next, we will draw some attention to performance across different graph types. The topology of NAS-Bench-\{101, 201, 301\} is similar in that they are DAGs where non-input nodes may have more than one edge feeding into them, and each non-output node may have more than one edge departing from them, but graphs have no more than 15 (usually less than 10) nodes. We see that in this context GEN, PNA, and 
GT tend to perform the best. %
In contrast, OFA-MBv3 and OFA-PN are sequence graphs where each node has at most one leading edge and one departing edge. Here, GEN and PNA continue to stand out. %
Finally, we have ONNX-IR where the graphs contain hundreds of nodes and many edges connecting them together. For hop-level SRCC prediction, after GEN, GIN, PNA, GATv2 and GT achieve high performance with several ties for 2nd and 3rd place. %

\subsection{Additional Hardware Results}

Figure~\ref{fig:supp_hw} illustrates an additional hardware comparison for PINAT and AutoBuild on NB-101. On PINAT the $k$-GNN is totally dominant as it achieves the best performance, lowest latency and has a small footprint using the PINAT predictor. %
For AutoBuild on NB-101 for %
average hop-level SRCC, %
we can see the most optimal GNN layer types are $k$-GNN, which minimizes latency, and GEN, which maximizes SRCC while not being too costly.

\subsubsection{Memory Usage and Parameter Counts.}

Table~\ref{tab:supp_mem} provides the peak memory usage statistics for FlowerFormer, PINAT and AutoBuild on NAS-Bench-101. These numbers provide the scale to the icon sizes in Figures~\ref{fig:hw} and \ref{fig:supp_hw}. Further, %
Table~\ref{tab:supp_params} lists the parameter counts of predictors for FlowerFormer, PINAT and AutoBuild on NAS-Bench-101.

\begin{table}[t!] 
    \centering 
    \scalebox{\scaleboxratio}{ 
    \begin{tabular}{l|cccccccccccc} \toprule 
    \textbf{Method} & \textbf{GCN} & \textbf{SAGE} & \textbf{GIN} & \textbf{GEN} & \textbf{$k$-GNN} & \textbf{PNA} & \textbf{GATv2-H1} & \textbf{GATv2-H4} & \textbf{GT-H1} & \textbf{GT-H4} & \textbf{DAG} & \textbf{GATSet} \\ \midrule \textbf{FlowerFormer} & 82.5 & 84.6 & 88.9 & 105.9 & 84.5 & 140.8 & 96.58 & 146.5 & 91.2 & 119.9 & 172.9 & -- \\ 
    \textbf{PINAT} & 26.7 & 27.5 & 26.6 & 33.2 & 27.5 & 39.3 & 27.6 & 35.3 & 29.1 & 36.2 & -- & 28.1 \\ 
    \textbf{AutoBuild} & 66.1 & 66.3 & 67.23 & 70.9 & 66.3 & 76.2 & 69.5 & 81.6 & 67.8 & 75.2 & -- & -- \\ 
    \bottomrule 
    \end{tabular} 
    } 
    \caption{Peak memory usage (MB) for Figs.~\ref{fig:hw} and \ref{fig:supp_hw}.} 
    \label{tab:supp_mem} 
\end{table}

\begin{table}[t!] 
    \centering 
    \scalebox{\scaleboxratio}{ 
    \begin{tabular}{l|ccccccccccccc} \toprule \textbf{Method} & \textbf{MLP} & \textbf{GCN} & \textbf{SAGE} & \textbf{GIN} & \textbf{GEN} & \textbf{$k$-GNN} & \textbf{PNA} & \textbf{GATv2-H1} & \textbf{GATv2-H4} & \textbf{GT-H1} & \textbf{GT-H4} & \textbf{DAG} & \textbf{GATSet} \\ \midrule \textbf{FlowerFormer} & 396k & 412k & 449k & 396k & 525k & 449k & 967k & 414k & 677k & 525k & 862k & 901k & -- \\ \textbf{PINAT} & 479k & 494k & 534k & 479k & 832k & 534k & 1.15M & 537k & 787k & 617k & 989k & -- & 554k \\ \textbf{AutoBuild} & 5.6k & 4.7k & 7.8k & 5.6k & 18.4k & 7.8k & 52.8k & 8.2k & 28.2k & 14.5k & 43.8k & -- & -- \\ \bottomrule 
    \end{tabular} 
    } 
    \caption{Comparing the parameter counts of GNN predictors for FlowerFormer, PINAT and AutoBuild on NB-101.} \label{tab:supp_params} %
    \end{table}

\subsubsection{Parameter-matching.}
To separate MP mechanism from model capacity, we re-train AutoBuild on OFA-MBv3, OFA-PN and ONNX-IR with each layer's embedding dimension adjusted so that its parameter count matches GEN's default budget. GEN itself is the reference and is unchanged. Results are provided in Table~\ref{tab:autobuild_compute_match_mae}. %
On the two sequence-graph datasets, conclusions largely hold: GraphSAGE and $k$-GNN remain the best layers on OFA-PN and are among the best on OFA-MBv3, where the MLP baseline achieves the lowest error. On the operation-level ONNX-IR graphs, by contrast, resizing increases both the error and the seed variance of every resized layer, whether widened (GIN, GraphSAGE, $k$-GNN) or narrowed (PNA, GATv2, GT), and GATv2-H4 diverges. %
GEN consequently attains the lowest ONNX-IR error, followed by both GT variants, which remain stable. Because this reordering arises from other layers degrading rather than from GEN improving, we attribute it to an interaction between model width and the repository's fixed training recipe rather than to a capacity advantage of GEN. %

\begin{table*}[t!]
    \centering
    \scalebox{\scaleboxratio}{
    \begin{tabular}{lcccccccccccc} \toprule
    \textbf{Dataset} & \textbf{MLP} & \textbf{GCN} & \textbf{SAGE} & \textbf{GIN} &  \textbf{GEN} & \textbf{$k$-GNN} & \textbf{PNA} & \textbf{GATv2-H1} & \textbf{GATv2-H4} & \textbf{GT-H1} & \textbf{GT-H4} \\ \midrule

OFA-MBv3 & $\mathbf{0.26_{0.01}}$ & $0.34_{0.01}$ & $\underline{0.30_{0.01}}$ & $0.38_{0.01}$ & $0.39_{0.08}$ & $\underline{0.30_{0.01}}$ & $0.32_{0.03}$ & $0.34_{0.02}$ & $0.36_{0.05}$ & $\mathit{0.31}_{0.02}$ & $0.32_{0.01}$ \\

OFA-PN & $0.21_{0.07}$ & $0.17_{0.01}$ & $\mathbf{0.11_{0.02}}$ & $0.14_{0.02}$ & $0.14_{0.01}$ & $\underline{0.11_{0.02}}$ & $0.19_{0.03}$ & $\mathit{0.12_{0.01}}$ & $0.15_{0.07}$ & $0.13_{0.03}$ & $0.13_{0.01}$ \\

ONNX-IR & $0.27_{0.00}$ & $0.34_{0.19}$ & $0.31_{0.18}$ & $0.27_{0.18}$ & $\mathbf{0.15}_{0.03}$ & $0.44_{0.17}$ & $0.42_{0.18}$ & $0.45_{0.36}$ & $3.20_{3.5}$ & $\underline{0.17_{0.02}}$ & $\mathit{0.19_{0.04}}$ \\

\bottomrule
    \end{tabular}
    }
    \caption{AutoBuild parameter-matched MAE Results, achieved by adjusting the node feature embedding dimension for all layers except GEN. 
    Horizontal lines demarcate different datasets. %
    Results averaged across 5 random seeds.}
    \label{tab:autobuild_compute_match_mae}
\end{table*}

\subsection{Statistical Significance}
\label{sec:stats}
We test whether rank differences between GNN layers generalize across benchmarks using a Friedman test followed by a post-hoc Nemenyi test. Since training splits and metrics within a single dataset are strongly correlated, we do not treat individual trials as independent. Instead, we collapse trials into blocks in two ways. In the first, each regressor-dataset pair forms one block ($N=16$), in which a GNN's ranks are averaged over that pair's training splits and primary metrics, i.e., from Tab.~\ref{tab:datasets}. %
In the second, each dataset forms one block ($N=11$); datasets evaluated by several regressors (NB-101, NB-201 and NB-301) average the corresponding regressor-dataset ranks with equal weight per regressor. In both cases, missing results (PNA on FlowerFormer NB-301, the MLP on SDXL) receive the worst rank, and only the 11 layers common to every regressor are compared.

\begin{figure}[t]
    \centering
    \begin{subfigure}{1.0\linewidth}
        \centering
        \includegraphics[width=0.9\linewidth]{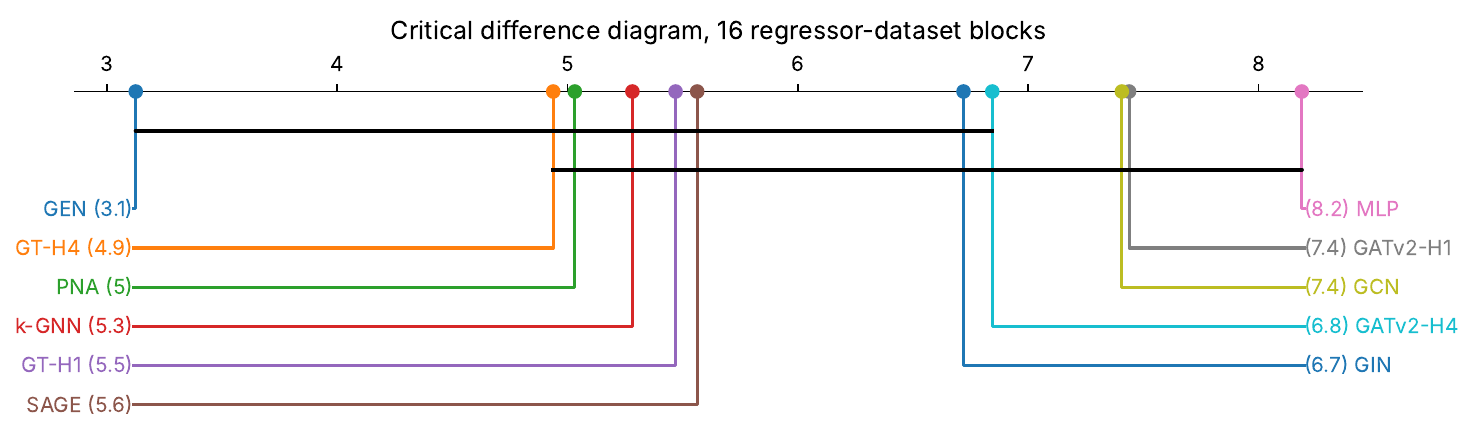}
        \caption{Regressor--dataset blocks ($N=16$; Friedman $\chi^2 = 31.41$, $p = 5.0 \times 10^{-4}$).}
        \label{fig:cd_test1}
    \end{subfigure}

    \vspace{2mm}

    \begin{subfigure}{1.0\linewidth}
        \centering
        \includegraphics[width=0.9\linewidth]{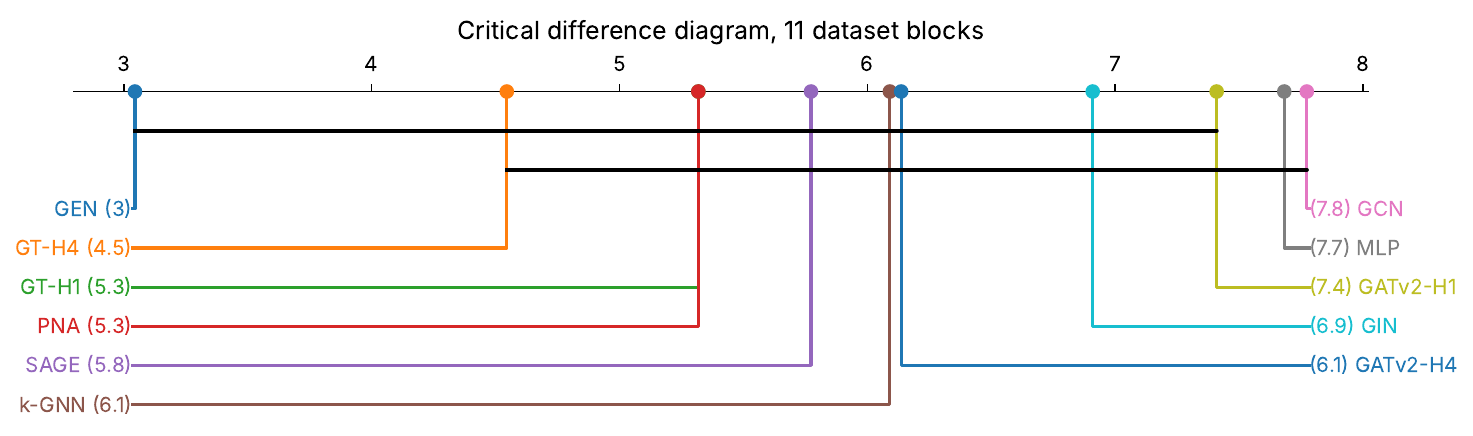}
        \caption{Dataset blocks ($N=11$; Friedman $\chi^2 = 20.83$, $p = 0.022$).}
        \label{fig:cd_test2}
    \end{subfigure}
    \caption{Critical difference diagrams of mean GNN rank (lower is better) on each repository's primary metrics. Each block averages a GNN's ranks over training splits and primary metrics; in (b), datasets used by several regressors weight each regressor equally. GNNs joined by a horizontal bar are not significantly different under the Nemenyi post-hoc test ($p \geq 0.05$). Missing results (PNA on FlowerFormer NB-301, MLP on SDXL) receive the worst rank.}
    \label{fig:cd_diagrams}
\end{figure}

\begin{figure}[t]
    \centering
    \begin{subfigure}{0.49\linewidth}
        \centering
        \includegraphics[width=\linewidth]{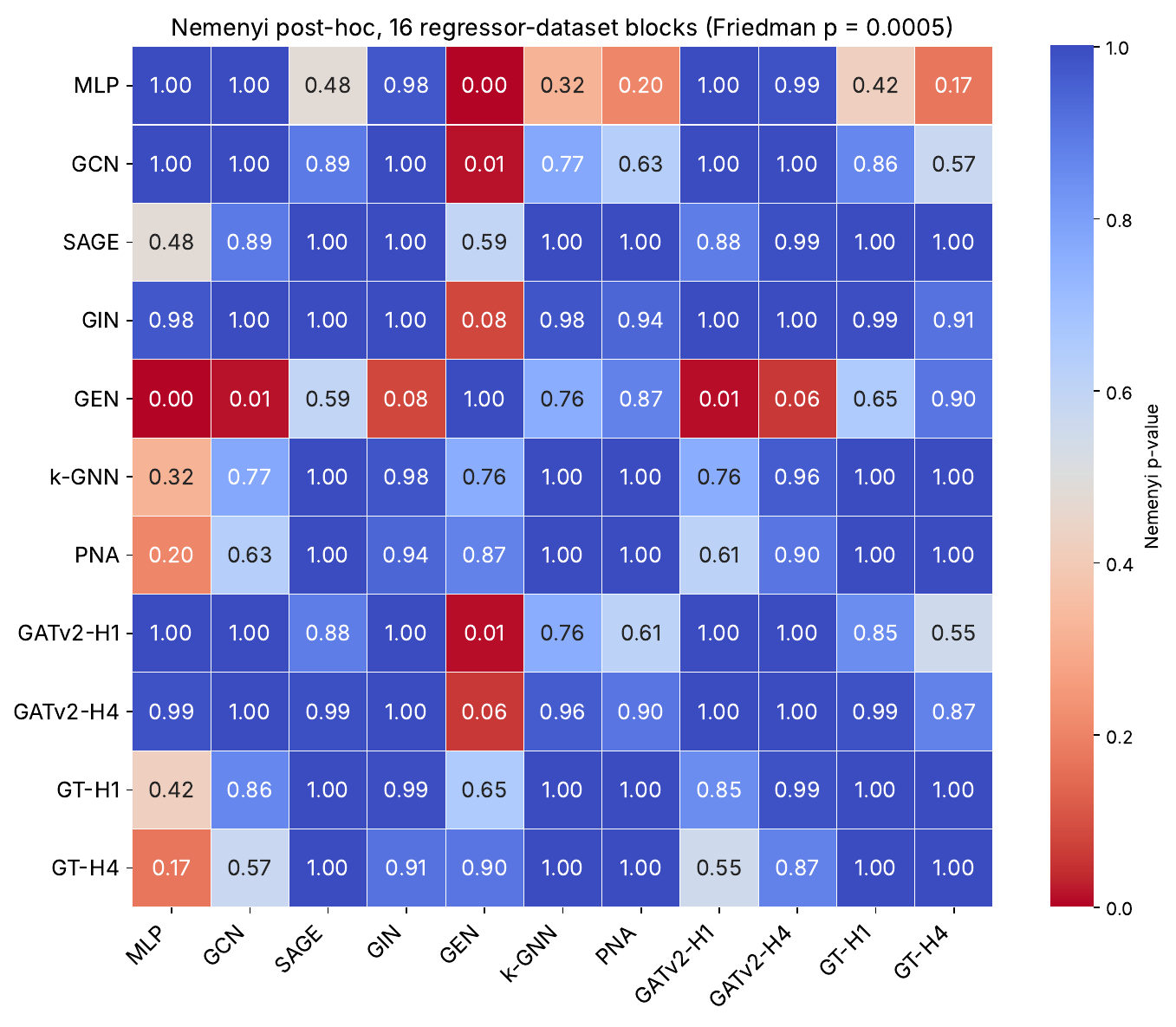}
        \caption{Regressor--dataset blocks ($N=16$).}
        \label{fig:nemenyi_test1}
    \end{subfigure}
    \hfill
    \begin{subfigure}{0.49\linewidth}
        \centering
        \includegraphics[width=\linewidth]{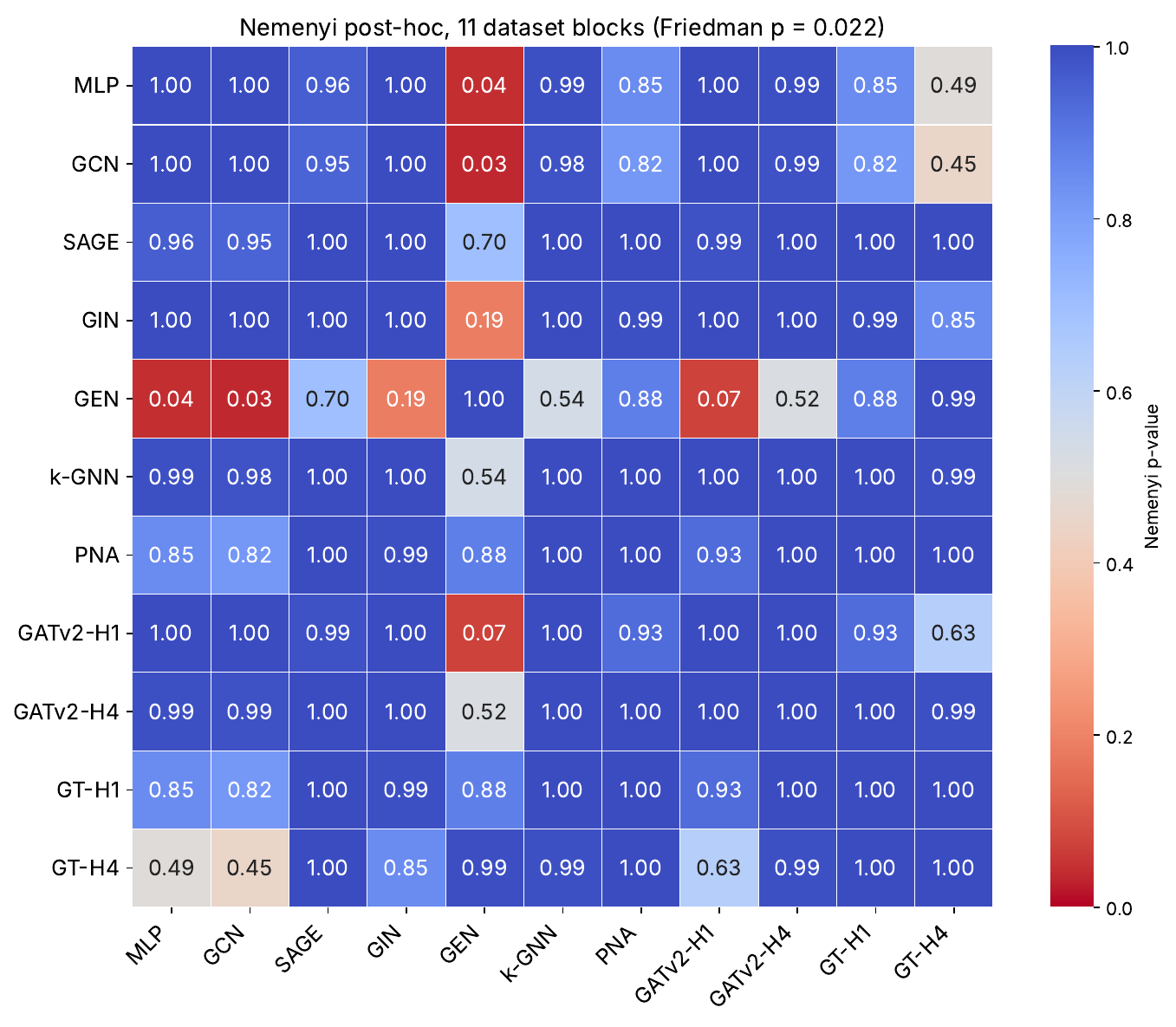}
        \caption{Dataset blocks ($N=11$).}
        \label{fig:nemenyi_test2}
    \end{subfigure}
    \caption{Pairwise Nemenyi post-hoc $p$-values following the Friedman tests of Fig.~\ref{fig:cd_diagrams}. Lower values indicate stronger evidence of a rank difference; pairs with $p \geq 0.05$ are not statistically distinguishable.}
    \label{fig:nemenyi_heatmaps}
\end{figure}

Both aggregations reject the null hypothesis of equal ranks (regressor-dataset blocks: $\chi^2 = 31.41$, $p = 5.0 \times 10^{-4}$; dataset blocks: $\chi^2 = 20.83$, $p = 0.022$). GEN attains the best mean rank under both (3.12 and 3.05), followed by GT-H4 (4.94 and 4.55). The Nemenyi test finds GEN significantly better than the MLP, GCN and GATv2-H1 with regressor--dataset blocks, and than the MLP and GCN with dataset blocks ($p < 0.05$). No other pair differs significantly, and GEN is not distinguishable from GT, PNA, $k$-GNN or GraphSAGE. Figure~\ref{fig:cd_diagrams} shows the critical difference diagrams and Figure~\ref{fig:nemenyi_heatmaps} the pairwise $p$-values. These results support our main conclusion: MP choice matters, but beyond the weakest layers no single layer is reliably best.

\subsection{Additional Experimental Results}

We enumerate additional results for each GNN regressor repository and supported dataset. Specifically, we include the primary and secondary evaluation metrics in this comparison. Just like the main body of this manuscript, for all tabular performance results, we report the \textbf{best}, \underline{second best} and \textit{third best} GNN MP mechanisms with \textbf{bold}, \underline{underlined} or %
\textit{italicized} text, respectively.

\subsubsection{FlowerFormer.}

Table~\ref{tab:supp_ff_nb101} enumerates our FlowerFormer results on NB-101 including KT, SRCC, MAE and R$^2$. We can see a few distinct patterns here: To contrast SRCC to KT, more GNN types are able to perform as well as DAGFormer at different training dataset levels when SRCC is measured. When considering MAE, the MLP baseline achieves the best performance at all data split levels. However, it is important to note that the FlowerFormer loss Eq.~\ref{eq:ff_loss} only aims to maximize ranking correlation, not minimize regression error. This is reflected in the R$^2$ metrics which are scattered, although the $k$-GNN and PNA do noticeably well. 

\begin{table*}[t!]
    \centering
    \scalebox{\scaleboxratio}{
    \begin{tabular}{lcccccccccccc} \toprule
    \textbf{NB-101} & \textbf{MLP} & \textbf{GCN} & \textbf{SAGE} & \textbf{GIN} &  \textbf{GEN} & \textbf{$k$-GNN} & \textbf{PNA} & \textbf{GATv2-H1} & \textbf{GATv2-H4} & \textbf{GT-H1} & \textbf{GT-H4} & \textbf{DAG} \\ \midrule
KT-1\% & $0.35_{0.04}$ & $0.72_{0.02}$ & $0.71_{0.03}$ & $0.68_{0.02}$ & $0.72_{0.02}$ & $\mathit{0.74}_{0.02}$& $\underline{0.75}_{0.03}$ & $0.72_{0.04}$ & $0.72_{0.02}$ & $\mathit{0.74}_{0.02}$ & $\mathit{0.74}_{0.01}$ & $\mathbf{0.78}_{0.02}$ \\
KT-5\% & $0.40_{0.01}$ & $\mathit{0.84}_{0.00}$ & $\underline{0.85}_{0.01}$ & $\mathit{0.84}_{0.01}$ & $\mathit{0.84}_{0.00}$ & $\underline{0.85}_{0.00}$ & $\underline{0.85}_{0.00}$ & $\mathit{0.84}_{0.01}$ & $\mathit{0.84}_{0.00}$ & $\mathit{0.84}_{0.00}$ & $\underline{0.85}_{0.00}$ & $\mathbf{0.86}_{0.01}$ \\
KT-10\% & $0.40_{0.02}$ & $\underline{0.87}_{0.00}$ & $\underline{0.87}_{0.00}$ & $\underline{0.87}_{0.00}$ & $\underline{0.87}_{0.00}$ & $\underline{0.87}_{0.00}$ & $\underline{0.87}_{0.00}$ & $\mathit{0.86}_{0.00}$ & $\mathit{0.86}_{0.00}$ & $\mathit{0.86}_{0.00}$ & $\underline{0.87}_{0.00}$ & $\mathbf{0.88}_{0.00}$ \\
KT-50\% & $\mathit{0.42}_{0.00}$ & $\mathbf{0.90}_{0.00}$ & $\mathbf{0.90}_{0.00}$ & $\underline{0.89}_{0.00}$ & $\underline{0.89}_{0.00}$ & $\mathbf{0.90}_{0.00}$ & $\underline{0.89}_{0.00}$ & $\underline{0.89}_{0.00}$ & $\underline{0.89}_{0.00}$ & $\underline{0.89}_{0.00}$ & $\underline{0.89}_{0.00}$ & $\underline{0.89}_{0.00}$ \\ \midrule

SRCC-1\% & $0.50_{0.06}$ & $0.89_{0.02}$ & $0.89_{0.02}$ & $0.87_{0.02}$ & $\mathit{0.90}_{0.01}$ & $\underline{0.91}_{0.01}$ & $\underline{0.91}_{0.02}$ & $\mathit{0.90}_{0.03}$ & $\mathit{0.90}_{0.02}$ & $\underline{0.91}_{0.01}$ & $\underline{0.91}_{0.01}$ & $\mathbf{0.94}_{0.01}$ \\
SRCC-5\% & $\mathit{0.57}_{0.02}$ & $\mathbf{0.97}_{0.00}$ & $\mathbf{0.97}_{0.00}$ & $\underline{0.96}_{0.00}$ &$\mathbf{0.97}_{0.00}$ & $\mathbf{0.97}_{0.00}$ & $\mathbf{0.97}_{0.00}$ & $\mathbf{0.97}_{0.00}$ & $\mathbf{0.97}_{0.00}$ & $\mathbf{0.97}_{0.00}$ & $\mathbf{0.97}_{0.00}$ & $\mathbf{0.97}_{0.00}$ \\
SRCC-10\% & $\mathit{0.57}_{0.02}$ & $\mathbf{0.98}_{0.00}$ & $\mathbf{0.98}_{0.00}$ & $\mathbf{0.98}_{0.00}$ & $\mathbf{0.98}_{0.00}$ & $\mathbf{0.98}_{0.00}$ & $\mathbf{0.98}_{0.00}$ & $\underline{0.97}_{0.00}$ & $\mathbf{0.98}_{0.00}$ & $\underline{0.97}_{0.00}$ & $\mathbf{0.98}_{0.00}$ & $\mathbf{0.98}_{0.00}$ \\
SRCC-50\% & $\mathit{0.58}_{0.00}$ & $\mathbf{0.99}_{0.00}$ & $\mathbf{0.99}_{0.00}$ & $\underline{0.98}_{0.00}$ & $\underline{0.98}_{0.00}$ & $\underline{0.98}_{0.00}$ & $\underline{0.98}_{0.00}$ & $\underline{0.98}_{0.00}$ & $\underline{0.98}_{0.00}$ & $\underline{0.98}_{0.00}$ & $\underline{0.98}_{0.00}$ & $\underline{0.98}_{0.00}$ \\ \midrule

MAE-1\% & $\mathbf{0.92}_{0.12}$ & $\underline{2.22}_{0.05}$ & $\mathit{2.30}_{0.16}$ & $2.62_{0.19}$ & $2.61_{0.18}$ & $2.49_{0.10}$ & $2.83_{0.20}$ & $2.42_{0.12}$ & $2.51_{0.11}$ & $2.44_{0.17}$ & $2.47_{0.13}$ & $2.63_{0.19}$ \\
MAE-5\% & $\mathbf{0.90}_{0.09}$ & $\underline{2.55}_{0.10}$ & $\mathit{2.68}_{0.07}$ & $3.35_{0.10}$ & $3.15_{0.05}$ & $3.06_{0.14}$ & $3.40_{0.11}$ & $3.02_{0.08}$ & $2.79_{0.07}$ & $3.27_{0.12}$ & $3.13_{0.12}$ & $3.50_{0.05}$ \\
MAE-10\% & $\mathbf{0.88}_{0.07}$ & $\underline{2.86}_{0.11}$ & $\mathit{2.99}_{0.07}$ & $3.88_{0.07}$ & $3.63_{0.06}$ & $3.38_{0.07}$ & $3.84_{0.10}$ & $3.50_{0.09}$ & $3.23_{0.11}$ & $3.70_{0.07}$ & $3.62_{0.08}$ & $4.02_{0.09}$ \\
MAE-50\% & $\mathbf{0.89}_{0.06}$ & $\mathit{3.52}_{0.07}$ & $\underline{3.45}_{0.03}$ & $4.95_{0.13}$ & $4.56_{0.07}$ & $3.95_{0.10}$ & $5.01_{0.07}$ & $4.66_{0.04}$ & $4.37_{0.07}$ & $4.58_{0.12}$ & $4.70_{0.13}$ & $6.30_{0.25}$ \\ \midrule

R$^2$-1\% & $0.07_{0.02}$ & $\mathit{0.21}_{0.01}$ & $0.20_{0.02}$ & $\mathit{0.21}_{0.02}$ & $\mathit{0.21}_{0.01}$ & $\mathbf{0.24}_{0.01}$ & $\underline{0.22}_{0.02}$ & $\mathit{0.21}_{0.02}$ & $0.20_{0.01}$ & $\underline{0.22}_{0.01}$ & $\underline{0.22}_{0.01}$ & $\underline{0.22}_{0.01}$ \\
R$^2$-5\% & $\mathit{0.11}_{0.00}$ & $\underline{0.24}_{0.01}$ & $\underline{0.24}_{0.01}$ & $\mathbf{0.25}_{0.01}$ & $\underline{0.24}_{0.01}$ & $\mathbf{0.25}_{0.01}$ & $\mathbf{0.25}_{0.02}$ & $\underline{0.24}_{0.01}$ &$\underline{0.24}_{0.01}$ & $\underline{0.24}_{0.00}$ & $\mathbf{0.25}_{0.00}$ & $\underline{0.24}_{0.02}$ \\
R$^2$-10\% & $0.11_{0.01}$ & $\mathit{0.24}_{0.00}$ & $\mathit{0.24}_{0.01}$ & $\mathbf{0.26}_{0.01}$ & $\underline{0.25}_{0.00}$ & $\underline{0.25}_{0.01}$ & $\mathbf{0.26}_{0.01}$ & $\mathit{0.24}_{0.01}$ & $\mathit{0.24}_{0.01}$ & $\underline{0.25}_{0.00}$ & $\underline{0.25}_{0.01}$ & $\mathbf{0.26}_{0.01}$ \\
R$^2$-50\% & $\mathit{0.11}_{0.00}$ & $\underline{0.26}_{0.00}$ & $\mathbf{0.27}_{0.01}$ & $\underline{0.26}_{0.01}$ & $\mathbf{0.27}_{0.01}$ & $\mathbf{0.27}_{0.01}$ & $\mathbf{0.27}_{0.01}$ & $\underline{0.26}_{0.01}$ & $\underline{0.26}_{0.00}$ & $\underline{0.26}_{0.00}$ & $\underline{0.26}_{0.00}$ & $\underline{0.26}_{0.01}$ \\\bottomrule
    \end{tabular}
    }
    \caption{FlowerFormer NAS-Bench-101 results. Rows correspond to different evaluation metrics and percentage proportion of the dataset utilized for training. For KT, SRCC and R$^2$, higher is better. For MAE, lower is better. Horizontal lines demarcate different classes of metrics. %
    Results averaged across 5 random seeds.}
    \label{tab:supp_ff_nb101}
\end{table*}

\begin{table*}[t!]
    \centering
    \scalebox{\scaleboxratio}{
    \begin{tabular}{lcccccccccccc} \toprule
    \textbf{NB-201} & \textbf{MLP} & \textbf{GCN} & \textbf{SAGE} & \textbf{GIN} &  \textbf{GEN} & \textbf{$k$-GNN} & \textbf{PNA} & \textbf{GATv2-H1} & \textbf{GATv2-H4} & \textbf{GT-H1} & \textbf{GT-H4} & \textbf{DAG} \\ \midrule
KT-1\% & $\mathit{0.71}_{0.04}$ & $\mathit{0.71}_{0.02}$ & $\underline{0.72}_{0.02}$ & $0.68_{0.03}$ & $\mathit{0.71}_{0.01}$ & $\underline{0.72}_{0.02}$ & $\mathit{0.71}_{0.02}$ & $0.69_{0.03}$ & $\mathit{0.71}_{0.03}$ & $0.67_{0.02}$ & $0.70_{0.01}$ & $\mathbf{0.77}_{0.01}$ \\
KT-5\% & $\mathit{0.85}_{0.01}$ & $\underline{0.86}_{0.01}$ & $\underline{0.86}_{0.01}$ & $0.84_{0.01}$ & $\underline{0.86}_{0.01}$ & $\underline{0.86}_{0.00}$ & $\underline{0.86}_{0.00}$ & $\mathit{0.85}_{0.01}$ & $\underline{0.86}_{0.00}$ & $0.83_{0.01}$ & $0.84_{0.00}$ & $\mathbf{0.89}_{0.01}$ \\
KT-10\% & $0.88_{0.00}$ & $\underline{0.90}_{0.00}$ & $\mathbf{0.91}_{0.00}$ & $\mathit{0.89}_{0.01}$ & $\underline{0.90}_{0.00}$ & $\underline{0.90}_{0.00}$ & $\underline{0.90}_{0.00}$ & $\mathit{0.89}_{0.00}$ & $\underline{0.90}_{0.00}$ & $0.88_{0.01}$ & $0.88_{0.00}$ & $\mathbf{0.91}_{0.00}$ \\
KT-50\% & $\underline{0.89}_{0.00}$ & $\mathbf{0.93}_{0.00}$ & $\mathbf{0.93}_{0.00}$ & $\mathbf{0.93}_{0.00}$ & $\mathbf{0.93}_{0.00}$ & $\mathbf{0.93}_{0.00}$ & $\mathbf{0.93}_{0.00}$ & $\mathbf{0.93}_{0.00}$ & $\mathbf{0.93}_{0.00}$ & $\mathbf{0.93}_{0.00}$ & $\mathbf{0.93}_{0.00}$ & $\mathbf{0.93}_{0.00}$ \\ \midrule

SRCC-1\% & $\mathit{0.88}_{0.03}$ & $\underline{0.89}_{0.02}$ & $\underline{0.89}_{0.01}$ & $0.86_{0.03}$ & $\underline{0.89}_{0.01}$ & $\underline{0.89}_{0.02}$ & $\underline{0.89}_{0.02}$ & $0.87_{0.03}$ & $\underline{0.89}_{0.02}$ & $0.85_{0.01}$ & $\mathit{0.88}_{0.00}$ & $\mathbf{0.93}_{0.01}$ \\
SRCC-5\% & $\underline{0.97}_{0.00}$ & $\underline{0.97}_{0.00}$ & $\underline{0.97}_{0.00}$ & $\mathit{0.96}_{0.01}$ & $\underline{0.97}_{0.00}$ & $\underline{0.97}_{0.00}$ & $\underline{0.97}_{0.00}$ & $\mathit{0.96}_{0.00}$ & $\underline{0.97}_{0.00}$ & $\mathit{0.96}_{0.00}$ & $\mathit{0.96}_{0.00}$ & $\mathbf{0.98}_{0.00}$ \\
SRCC-10\% & $\underline{0.98}_{0.00}$ & $\mathbf{0.99}_{0.00}$ & $\mathbf{0.99}_{0.00}$ & $\underline{0.98}_{0.00}$ & $\mathbf{0.99}_{0.00}$ & $\mathbf{0.99}_{0.00}$ & $\mathbf{0.99}_{0.00}$ & $\underline{0.98}_{0.00}$ & $\underline{0.98}_{0.00}$ & $\underline{0.98}_{0.00}$ & $\underline{0.98}_{0.00}$ & $\mathbf{0.99}_{0.00}$ \\
SRCC-50\% & $\underline{0.98}_{0.00}$ & $\mathbf{0.99}_{0.00}$ & $\mathbf{0.99}_{0.00}$ & $\mathbf{0.99}_{0.00}$ & $\mathbf{0.99}_{0.00}$ & $\mathbf{0.99}_{0.00}$ & $\mathbf{0.99}_{0.00}$ & $\mathbf{0.99}_{0.00}$ & $\mathbf{0.99}_{0.00}$ & $\mathbf{0.99}_{0.00}$ & $\mathbf{0.99}_{0.00}$ & $\mathbf{0.99}_{0.00}$ \\ \midrule

MAE-1\% & $\underline{3.07}_{0.21}$ & $\mathbf{3.05}_{0.19}$ & $\mathit{3.26}_{0.19}$ & $4.04_{0.25}$ & $3.66_{0.25}$ & $3.37_{0.15}$ & $3.78_{0.17}$ & $3.61_{0.14}$ & $3.38_{0.23}$ & $3.82_{0.03}$ & $3.79_{0.22}$ & $4.42_{0.22}$ \\
MAE-5\% & $\mathbf{3.91}_{0.13}$ & $\mathit{4.69}_{0.09}$ & $\underline{4.52}_{0.29}$ & $6.24_{0.13}$ & $5.87_{0.13}$ & $5.03_{0.23}$ & $5.70_{0.15}$ & $5.81_{0.08}$ & $5.49_{0.22}$ & $5.62_{0.21}$ & $5.93_{0.15}$ & $9.17_{0.33}$ \\
MAE-10\% & $\mathbf{4.73}_{0.16}$ & $\mathit{6.67}_{0.11}$ & $\underline{6.46}_{0.27}$ & $8.28_{0.03}$ & $8.55_{0.19}$ & $7.32_{0.09}$ & $8.82_{0.23}$ & $8.26_{0.20}$ & $8.75_{0.19}$ & $8.15_{0.32}$ & $8.73_{0.14}$ & $17.20_{0.39}$ \\
MAE-50\% & $\mathbf{3.79}_{0.11}$ & $\mathit{8.90}_{0.51}$ & $\underline{7.51}_{0.25}$ & $12.17_{0.39}$ & $12.59_{0.41}$ & $9.01_{0.13}$ & $12.68_{0.62}$ & $11.03_{0.22}$ & $14.04_{0.52}$ & $9.25_{0.51}$ & $10.41_{0.50}$ & $33.14_{1.01}$ \\ \midrule

R$^2$-1\% & $\underline{0.30}_{0.05}$ & $\mathit{0.28}_{0.05}$ & $0.27_{0.03}$ & $0.22_{0.04}$ & $0.27_{0.03}$ & $0.26_{0.05}$ & $0.26_{0.05}$ & $0.24_{0.03}$ & $0.26_{0.03}$ & $0.23_{0.03}$ & $0.24_{0.02}$ & $\mathbf{0.32}_{0.01}$ \\
R$^2$-5\% & $\mathbf{0.40}_{0.01}$ & $\mathbf{0.40}_{0.02}$ & $\underline{0.39}_{0.02}$ & $0.35_{0.03}$ & $0.36_{0.02}$ & $0.37_{0.02}$ & $\mathit{0.38}_{0.02}$ & $\mathit{0.38}_{0.01}$ & $\underline{0.39}_{0.02}$ & $0.34_{0.02}$ & $0.36_{0.01}$ & $0.37_{0.01}$ \\
R$^2$-10\% & $\mathbf{0.45}_{0.02}$ & $\underline{0.42}_{0.01}$ & $\mathit{0.41}_{0.02}$ & $0.40_{0.01}$ & $0.39_{0.01}$ & $0.40_{0.01}$ & $0.40_{0.01}$ & $0.40_{0.02}$ & $\mathit{0.41}_{0.02}$ & $0.39_{0.02}$ & $0.39_{0.02}$ & $0.37_{0.01}$ \\
R$^2$-50\% & $\mathbf{0.56}_{0.02}$ & $\mathit{0.47}_{0.01}$ & $\underline{0.48}_{0.01}$ & $0.46_{0.01}$ & $0.44_{0.01}$ & $0.46_{0.02}$ & $0.46_{0.02}$ & $0.46_{0.01}$ & $0.45_{0.01}$ & $0.46_{0.02}$ & $0.44_{0.01}$ & $0.40_{0.01}$ \\ \bottomrule
    \end{tabular}
    }
    \caption{FlowerFormer NAS-Bench-201 Results. Same experimental setup as Tab.~\ref{tab:supp_ff_nb101}. }
    \label{tab:supp_ff_nb201}

\end{table*}

\begin{table*}[t!]
    \centering
    \scalebox{\scaleboxratio}{
    \begin{tabular}{lccccccccccc} \toprule
    \textbf{NB-301} & \textbf{MLP} & \textbf{GCN} & \textbf{SAGE} & \textbf{GIN} &  \textbf{GEN} & \textbf{$k$-GNN} & \textbf{GATv2-H1} & \textbf{GATv2-H4} & \textbf{GT-H1} & \textbf{GT-H4} & \textbf{DAG} \\ \midrule
KT-1\% & $0.50_{0.14}$ & $0.60_{0.03}$ & $\underline{0.62}_{0.01}$ & $0.59_{0.01}$ & $0.49_{0.21}$ & $\mathbf{0.63}_{0.02}$ & $\underline{0.62}_{0.01}$ & $0.54_{0.12}$ & $0.32_{0.17}$ & $0.32_{0.17}$ & $\mathit{0.61}_{0.02}$ \\
KT-5\% & $0.16_{0.12}$ & $0.22_{0.09}$ & $0.14_{0.10}$ & $0.22_{0.08}$ & $\mathbf{0.37}_{0.08}$ & $\underline{0.35}_{0.08}$ & $0.17_{0.11}$ & $0.20_{0.09}$ & $0.22_{0.05}$ & $0.19_{0.04}$ & $\mathit{0.26}_{0.02}$ \\
KT-10\% & $0.06_{0.14}$ & $0.26_{0.10}$ & $0.25_{0.10}$ & $\mathbf{0.37}_{0.07}$ & $\underline{0.34}_{0.10}$ & $\underline{0.34}_{0.07}$ & $0.02_{0.06}$ & $0.17_{0.13}$ & $0.27_{0.08}$ & $\mathit{0.30}_{0.05}$ & $0.28_{0.06}$ \\
KT-50\% & $0.22_{0.04}$ & $0.36_{0.08}$ & $0.36_{0.09}$ & $\underline{0.39}_{0.09}$ & $\mathit{0.37}_{0.06}$ & $\mathbf{0.59}_{0.05}$ & $0.02_{0.17}$ & $0.21_{0.09}$ & $0.28_{0.04}$ & $0.31_{0.06}$ & $0.28_{0.08}$ \\ \midrule

SRCC-1\% & $0.67_{0.17}$ & $0.78_{0.03}$ & $\underline{0.81}_{0.01}$ & $0.78_{0.01}$ & $0.65_{0.27}$ & $\mathbf{0.82}_{0.02}$ & $\underline{0.81}_{0.01}$ & $0.72_{0.15}$ & $0.45_{0.22}$ & $0.45_{0.22}$ & $\mathit{0.79}_{0.02}$ \\
SRCC-5\% & $0.23_{0.18}$ & $0.31_{0.13}$ & $0.21_{0.14}$ & $0.32_{0.12}$ & $\mathbf{0.52}_{0.09}$ & $\underline{0.50}_{0.10}$ & $0.24_{0.17}$ & $0.30_{0.13}$ & $0.32_{0.08}$ & $0.29_{0.06}$ & $\mathit{0.37}_{0.03}$ \\
SRCC-10\% & $0.09_{0.20}$ & $0.37_{0.13}$ & $0.36_{0.14}$ & $\mathbf{0.52}_{0.09}$ & $\underline{0.48}_{0.13}$ & $\underline{0.48}_{0.09}$ & $0.03_{0.09}$ & $0.24_{0.18}$ & $0.39_{0.11}$ & $\mathit{0.43}_{0.06}$ & $0.41_{0.08}$ \\
SRCC-50\% & $0.32_{0.05}$ & $0.51_{0.11}$ & $0.51_{0.12}$ & $\underline{0.55}_{0.12}$ & $\mathit{0.53}_{0.08}$ & $\mathbf{0.77}_{0.04}$ & $0.04_{0.25}$ & $0.31_{0.12}$ & $0.41_{0.05}$ & $0.45_{0.08}$ & $0.41_{0.11}$ \\ \midrule

MAE-1\% & $\mathit{1.03}_{0.13}$ & $1.24_{0.06}$ & $1.09_{0.13}$ & $1.51_{0.13}$ & $1.07_{0.20}$ & $1.24_{0.13}$& $1.33_{0.06}$ & $1.10_{0.17}$ & $\mathbf{0.98}_{0.15}$ & $\underline{0.99}_{0.16}$ & $1.70_{0.15}$ \\
MAE-5\% & $\underline{0.86}_{0.06}$ & $0.88_{0.03}$ & $\mathit{0.87}_{0.05}$ & $0.91_{0.07}$ & $0.88_{0.06}$ & $0.93_{0.05}$ & $\mathbf{0.84}_{0.07}$ & $0.90_{0.07}$ & $\mathbf{0.84}_{0.07}$ & $0.90_{0.05}$ & $0.91_{0.01}$ \\
MAE-10\% & $\underline{0.91}_{0.07}$ & $\mathit{0.92}_{0.06}$ & $\mathit{0.92}_{0.06}$ & $\underline{0.91}_{0.04}$ & $0.94_{0.04}$ & $0.95_{0.04}$ & $\mathbf{0.89}_{0.07}$ & $\mathbf{0.89}_{0.07}$ & $\mathit{0.92}_{0.06}$& $0.95_{0.07}$ & $0.93_{0.03}$ \\
MAE-50\% & $0.91_{0.09}$ & $0.92_{0.04}$ & $0.91_{0.07}$ & $\mathit{0.90}_{0.05}$ & $0.91_{0.05}$ & $\underline{0.86}_{0.05}$ & $\mathbf{0.79}_{0.12}$ & $\underline{0.86}_{0.08}$ & $\mathit{0.90}_{0.01}$ & $0.95_{0.06}$ &$0.98_{0.05}$ \\ \midrule

R$^2$-1\% & $0.27_{0.13}$ & $0.30_{0.06}$ & $\mathit{0.33}_{0.05}$ & $0.29_{0.02}$ & $0.23_{0.12}$ & $\mathbf{0.37}_{0.04}$ & $\underline{0.34}_{0.04}$ & $0.29_{0.08}$ & $0.13_{0.14}$ & $0.12_{0.12}$ & $0.29_{0.06}$ \\
R$^2$-5\% & $0.05_{0.03}$ & $\mathit{0.07}_{0.04}$ & $0.02_{0.02}$ & $0.05_{0.03}$ & $\mathbf{0.11}_{0.06}$ & $\underline{0.10}_{0.06}$ & $0.05_{0.05}$ & $0.04_{0.04}$ & $0.04_{0.02}$ & $0.03_{0.01}$ & $0.06_{0.01}$ \\
R$^2$-10\% & $0.03_{0.04}$ & $0.09_{0.07}$ & $0.07_{0.04}$ & $\mathbf{0.13}_{0.06}$ & $\underline{0.12}_{0.08}$ & $\mathit{0.11}_{0.04}$ & $0.00_{0.00}$ & $0.04_{0.04}$ & $0.07_{0.05}$ & $0.07_{0.02}$ & $0.08_{0.02}$ \\
R$^2$-50\% & $0.05_{0.03}$ & $0.12_{0.07}$ & $\mathit{0.13}_{0.04}$ & $\underline{0.15}_{0.05}$ & $\mathit{0.13}_{0.05}$ & $\mathbf{0.30}_{0.04}$ & $0.02_{0.02}$ & $0.05_{0.04}$ & $0.07_{0.01}$ & $0.09_{0.04}$ & $0.08_{0.04}$ \\ \bottomrule

    \end{tabular}
    }
    \caption{FlowerFormer NAS-Bench-301 Results. Same experimental setup as Tab.~\ref{tab:supp_ff_nb101}. PNA not included as it causes CUDA OOM on this dataset.}
    \label{tab:supp_ff_nb301}

\end{table*}

Next, Table~\ref{tab:supp_ff_nb201} enumerates results for FlowerFormer on NB-201. The results are quite similar to that of NB-101: when considering SRCC, more GNNs can match the performance of DAGFormer; when considering MAE, the MLP baseline performs very well. Unlike the NB-101 results, the MLP baseline does well on R$^2$ here as well. 

\begin{table*}[t!]
    \centering
    \scalebox{\scaleboxratio}{
    \begin{tabular}{lcccccccccccc} \toprule
    \textbf{NB-ASR} & \textbf{MLP} & \textbf{GCN} & \textbf{SAGE} & \textbf{GIN} &  \textbf{GEN} & \textbf{$k$-GNN} & \textbf{PNA} & \textbf{GATv2-H1} & \textbf{GATv2-H4} & \textbf{GT-H1} & \textbf{GT-H4} & \textbf{DAG} \\ \midrule
KT-1\% & $\underline{0.32}_{0.04}$ & $0.30_{0.07}$ & $\mathit{0.31}_{0.03}$ & $\underline{0.32}_{0.02}$ & $0.29_{0.05}$ & $0.28_{0.04}$ & $0.29_{0.06}$ & $\mathbf{0.33}_{0.04}$ & $\mathbf{0.33}_{0.04}$ & $0.29_{0.03}$ & $0.29_{0.03}$ & $0.24_{0.07}$ \\
KT-5\% & $\mathbf{0.46}_{0.02}$ & $\mathit{0.42}_{0.01}$ & $\underline{0.43}_{0.01}$ & $0.39_{0.01}$ & $\mathit{0.42}_{0.02}$ & $0.40_{0.02}$ & $0.41_{0.02}$ & $\mathit{0.42}_{0.01}$ & $\mathit{0.42}_{0.01}$ & $0.40_{0.02}$ & $0.40_{0.02}$ & $0.41_{0.03}$ \\
KT-10\% & $\mathbf{0.49}_{0.01}$ & $0.46_{0.01}$ & $\mathit{0.47}_{0.02}$ & $0.44_{0.01}$ & $\mathit{0.47}_{0.02}$ & $0.45_{0.01}$ & $\mathit{0.47}_{0.02}$ & $0.46_{0.01}$ & $0.46_{0.02}$ & $\mathit{0.47}_{0.02}$ & $0.46_{0.02}$ & $\underline{0.48}_{0.01}$ \\
KT-50\% & $\mathbf{0.59}_{0.00}$ & $0.52_{0.01}$ & $\mathit{0.54}_{0.02}$ & $0.51_{0.01}$ & $0.52_{0.01}$ & $0.52_{0.01}$ & $\underline{0.58}_{0.01}$ & $0.53_{0.03}$ & $0.52_{0.00}$ & $\underline{0.58}_{0.01}$ & $\underline{0.58}_{0.01}$ & $0.53_{0.01}$ \\ \midrule

SRCC-1\% & $\underline{0.46}_{0.05}$ & $0.43_{0.10}$ & $\mathit{0.44}_{0.04}$ & $\underline{0.46}_{0.03}$ & $0.42_{0.07}$ & $0.41_{0.06}$ & $0.42_{0.08}$ & $\mathbf{0.47}_{0.06}$ & $\mathbf{0.47}_{0.06}$ & $0.43_{0.04}$ & $0.42_{0.04}$ & $0.35_{0.10}$ \\
SRCC-5\% & $\mathbf{0.63}_{0.03}$ & $\mathit{0.58}_{0.01}$ & $\underline{0.59}_{0.02}$ & $0.55_{0.02}$ & $\mathit{0.58}_{0.02}$ & $0.56_{0.02}$ & $0.57_{0.03}$ & $\mathit{0.58}_{0.01}$ & $\mathit{0.58}_{0.02}$ & $0.56_{0.02}$ & $0.56_{0.02}$ & $0.57_{0.03}$ \\
SRCC-10\% & $\mathbf{0.66}_{0.01}$ & $\mathit{0.63}_{0.01}$ & $\mathit{0.63}_{0.02}$ & $0.60_{0.01}$ & $\mathit{0.63}_{0.02}$ & $0.61_{0.01}$ & $\mathit{0.63}_{0.02}$ & $0.62_{0.01}$ & $\mathit{0.63}_{0.02}$ & $\mathit{0.63}_{0.02}$ & $0.62_{0.03}$ & $\underline{0.64}_{0.02}$ \\
SRCC-50\% & $\mathbf{0.77}_{0.01}$ & $0.68_{0.01}$ & $\mathit{0.70}_{0.02}$ & $0.67_{0.01}$ & $0.68_{0.01}$ & $0.68_{0.01}$ & $\underline{0.76}_{0.01}$ & $\mathit{0.70}_{0.04}$ & $0.68_{0.01}$ & $\underline{0.76}_{0.01}$ &$\mathbf{0.77}_{0.01}$ & $0.69_{0.01}$ \\ \midrule

MAE-1\% & $\mathbf{0.77}_{0.08}$ & $0.93_{0.14}$ & $\mathit{0.86}_{0.09}$ & $0.88_{0.14}$ & $0.98_{0.14}$ & $\underline{0.82}_{0.12}$ & $0.95_{0.13}$ & $1.05_{0.16}$ & $0.98_{0.17}$ & $0.93_{0.10}$ & $\mathit{0.86}_{0.14}$ & $0.91_{0.10}$ \\
MAE-5\% & $\mathbf{0.62}_{0.10}$ & $0.89_{0.23}$ & $\mathit{0.84}_{0.18}$ & $1.10_{0.24}$ & $\underline{0.80}_{0.35}$ & $1.02_{0.23}$ & $\mathit{0.84}_{0.35}$ & $1.10_{0.15}$ & $1.22_{0.18}$ & $0.93_{0.39}$ & $1.00_{0.24}$ & $0.93_{0.35}$ \\
MAE-10\% & $\underline{0.48}_{0.06}$ & $0.65_{0.21}$ & $0.74_{0.11}$ & $0.66_{0.17}$ & $\mathbf{0.45}_{0.11}$ & $0.75_{0.29}$ & $0.76_{0.39}$ & $\mathit{0.61}_{0.06}$ & $0.87_{0.25}$ & $0.66_{0.16}$ & $0.78_{0.29}$ & $0.73_{0.19}$ \\
MAE-50\% & $0.81_{0.27}$ & $0.69_{0.09}$ & $1.04_{0.15}$ & $\mathbf{0.36}_{0.06}$ & $0.71_{0.29}$ & $\underline{0.39}_{0.09}$ & $0.89_{0.37}$ & $0.66_{0.19}$ & $0.79_{0.18}$ & $0.74_{0.41}$ & $0.87_{0.16}$ & $\mathit{0.41}_{0.08}$ \\ \midrule

R$^2$-1\% & $0.17_{0.05}$ & $\mathbf{0.25}_{0.10}$ & $0.12_{0.07}$ & $0.15_{0.06}$ & $0.17_{0.08}$ & $0.15_{0.05}$& $0.15_{0.08}$ & $\underline{0.23}_{0.06}$ & $0.18_{0.07}$ & $\mathit{0.19}_{0.06}$ & $0.18_{0.04}$ & $0.13_{0.10}$ \\
R$^2$-5\% & $\mathbf{0.41}_{0.07}$ & $0.38_{0.02}$ & $\underline{0.40}_{0.06}$ & $0.27_{0.05}$ & $0.37_{0.09}$ & $0.30_{0.06}$ & $0.31_{0.05}$ & $0.35_{0.02}$ & $0.31_{0.04}$ & $0.28_{0.07}$ & $\mathit{0.39}_{0.06}$ & $0.31_{0.11}$ \\
R$^2$-10\% & $\mathit{0.47}_{0.03}$ & $0.38_{0.03}$ & $\mathbf{0.51}_{0.05}$ & $0.37_{0.07}$ & $0.38_{0.05}$ & $0.44_{0.05}$ & $0.38_{0.04}$ & $0.43_{0.07}$ & $0.46_{0.02}$ & $0.39_{0.08}$ & $0.39_{0.11}$ & $\underline{0.49}_{0.10}$ \\
R$^2$-50\% & $\mathbf{0.54}_{0.02}$ & $0.45_{0.02}$ & $0.47_{0.04}$ & $0.47_{0.04}$ & $0.45_{0.03}$ & $0.46_{0.01}$ & $\mathit{0.48}_{0.01}$ & $\mathit{0.48}_{0.04}$ & $0.47_{0.02}$ & $\mathit{0.48}_{0.02}$ & $0.47_{0.01}$ & $\underline{0.49}_{0.05}$ \\ \bottomrule

    \end{tabular}
    }
    \caption{FlowerFormer NAS-Bench-ASR Results. Same experimental setup as Tab.~\ref{tab:supp_ff_nb101}. }
    \label{tab:supp_ff_nbasr}
\end{table*}
\begin{table*}[t!]
    \centering
    \scalebox{\scaleboxratio}{
    \begin{tabular}{lcccccccccccc} \toprule
    \textbf{NB-Graph} & \textbf{MLP} & \textbf{GCN} & \textbf{SAGE} & \textbf{GIN} &  \textbf{GEN} & \textbf{$k$-GNN} & \textbf{PNA} & \textbf{GATv2-H1} & \textbf{GATv2-H4} & \textbf{GT-H1} & \textbf{GT-H4} & \textbf{DAG} \\ \midrule
KT-1\% & $0.41_{0.02}$ & $\mathit{0.46}_{0.02}$ & $0.42_{0.03}$ & $0.45_{0.01}$ & $\mathit{0.46}_{0.01}$ & $0.45_{0.01}$ & $\underline{0.48}_{0.02}$ & $0.42_{0.02}$ & $0.44_{0.03}$ & $0.42_{0.01}$ & $0.44_{0.05}$ & $\mathbf{0.51}_{0.01}$ \\
KT-5\% & $0.55_{0.01}$ & $\mathit{0.61}_{0.01}$ & $0.60_{0.01}$ & $0.60_{0.01}$ & $\mathit{0.61}_{0.01}$ & $0.60_{0.01}$ & $\underline{0.62}_{0.01}$ & $0.59_{0.01}$ & $\mathit{0.61}_{0.01}$ & $0.59_{0.03}$ & $0.59_{0.02}$ & $\mathbf{0.65}_{0.01}$ \\
KT-10\% & $0.59_{0.01}$ & $\mathit{0.66}_{0.01}$ & $\mathit{0.66}_{0.01}$ & $0.65_{0.01}$ & $\mathit{0.66}_{0.01}$ & $\mathit{0.66}_{0.00}$ & $\underline{0.67}_{0.01}$ & $\mathit{0.66}_{0.00}$ & $\mathit{0.66}_{0.01}$ & $\mathit{0.66}_{0.01}$ & $\mathit{0.66}_{0.01}$ & $\mathbf{0.69}_{0.00}$ \\
KT-50\% & $0.65_{0.00}$ & $\mathbf{0.74}_{0.00}$ & $\mathbf{0.74}_{0.00}$ & $\underline{0.73}_{0.00}$ & $\mathbf{0.74}_{0.00}$ & $\mathbf{0.74}_{0.00}$ & $\mathit{0.66}_{0.16}$ & $\mathbf{0.74}_{0.00}$ & $\mathbf{0.74}_{0.00}$ & $\mathbf{0.74}_{0.00}$ & $\mathbf{0.74}_{0.00}$ & $\underline{0.73}_{0.00}$ \\ \midrule

SRCC-1\% & $0.58_{0.03}$ & $\mathit{0.65}_{0.03}$ & $0.59_{0.04}$ & $0.63_{0.01}$ & $\mathit{0.65}_{0.01}$ & $0.63_{0.01}$ & $\underline{0.66}_{0.03}$ & $0.59_{0.03}$ & $0.61_{0.04}$ & $0.59_{0.01}$ & $0.62_{0.06}$ & $\mathbf{0.69}_{0.02}$ \\
SRCC-5\% & $0.74_{0.01}$ & $\mathit{0.80}_{0.01}$ & $0.79_{0.01}$ & $\mathit{0.80}_{0.01}$ & $\mathit{0.80}_{0.01}$ & $\mathit{0.80}_{0.01}$ & $\underline{0.81}_{0.01}$ & $0.78_{0.01}$ & $\mathit{0.80}_{0.01}$ & $0.78_{0.02}$ & $0.78_{0.02}$ & $\mathbf{0.83}_{0.01}$ \\
SRCC-10\% & $0.78_{0.01}$ & $\mathit{0.85}_{0.01}$ & $\mathit{0.85}_{0.01}$ & $0.84_{0.01}$ & $\mathit{0.85}_{0.01}$ & $0.84_{0.00}$ & $\underline{0.86}_{0.01}$ & $0.84_{0.00}$ & $0.84_{0.01}$ & $\mathit{0.85}_{0.01}$ & $\mathit{0.85}_{0.00}$ & $\mathbf{0.87}_{0.00}$ \\
SRCC-50\% & $0.84_{0.00}$ & $\mathbf{0.91}_{0.00}$ & $\underline{0.90}_{0.00}$ & $\mathit{0.89}_{0.00}$ & $\underline{0.90}_{0.00}$ & $\underline{0.90}_{0.00}$ & $0.82_{0.16}$ & $\underline{0.90}_{0.00}$ & $\underline{0.90}_{0.00}$ & $\underline{0.90}_{0.00}$ & $\mathbf{0.91}_{0.00}$ & $\mathit{0.89}_{0.00}$ \\ \midrule

MAE-1\% & $\mathit{0.76}_{0.03}$ & $0.77_{0.05}$ & $\mathbf{0.73}_{0.05}$ & $\underline{0.75}_{0.02}$ & $0.78_{0.03}$ & $0.77_{0.03}$ & $\mathit{0.76}_{0.03}$ & $\mathit{0.76}_{0.04}$ & $\mathit{0.76}_{0.04}$ & $\underline{0.75}_{0.05}$ & $\mathit{0.76}_{0.04}$ & $\mathbf{0.73}_{0.03}$ \\
MAE-5\% & $\mathit{0.75}_{0.04}$ & $0.77_{0.04}$ & $\underline{0.73}_{0.04}$ & $\mathit{0.75}_{0.02}$ & $0.78_{0.04}$ & $\mathit{0.75}_{0.02}$ & $0.76_{0.03}$ & $0.76_{0.04}$ & $\mathit{0.75}_{0.04}$ & $\mathit{0.75}_{0.05}$ & $0.77_{0.05}$ & $\mathbf{0.72}_{0.03}$ \\
MAE-10\% & $\underline{0.74}_{0.05}$ & $0.78_{0.05}$ & $\underline{0.74}_{0.05}$ & $\mathit{0.75}_{0.03}$ & $0.79_{0.04}$ & $0.77_{0.02}$ & $0.76_{0.03}$ & $0.76_{0.04}$ & $\mathit{0.75}_{0.05}$ & $0.76_{0.04}$ & $0.76_{0.04}$ & $\mathbf{0.72}_{0.03}$ \\
MAE-50\% & $0.76_{0.04}$ & $0.77_{0.05}$ & $\mathit{0.75}_{0.04}$ & $0.76_{0.02}$ & $0.79_{0.04}$ & $0.77_{0.03}$ & $\mathit{0.75}_{0.03}$ & $\mathit{0.75}_{0.04}$ & $\mathit{0.75}_{0.04}$ & $\underline{0.74}_{0.04}$ & $0.76_{0.05}$ & $\mathbf{0.72}_{0.03}$ \\ \midrule

R$^2$-1\% & $0.14_{0.01}$ & $0.19_{0.04}$ & $0.15_{0.03}$ & $\underline{0.23}_{0.03}$ & $0.17_{0.02}$ & $0.19_{0.03}$ & $\mathit{0.21}_{0.01}$ & $0.17_{0.02}$ & $0.16_{0.03}$ & $0.16_{0.02}$ & $0.18_{0.03}$ & $\mathbf{0.27}_{0.02}$ \\
R$^2$-5\% & $0.22_{0.01}$ & $\underline{0.34}_{0.01}$ & $0.31_{0.03}$ & $\underline{0.34}_{0.01}$ & $0.30_{0.01}$ & $\mathit{0.32}_{0.01}$ & $\mathit{0.32}_{0.02}$ & $0.30_{0.02}$ & $0.31_{0.02}$ & $0.30_{0.04}$ & $0.28_{0.02}$ & $\mathbf{0.36}_{0.01}$ \\
R$^2$-10\% & $0.26_{0.01}$ & $\mathbf{0.39}_{0.01}$ & $\mathbf{0.39}_{0.01}$ & $\underline{0.38}_{0.01}$ & $\mathit{0.37}_{0.01}$ & $\underline{0.38}_{0.01}$ & $\underline{0.38}_{0.03}$ & $\mathit{0.37}_{0.02}$ & $0.35_{0.01}$ & $\mathbf{0.39}_{0.01}$ & $\mathit{0.37}_{0.01}$ & $\mathbf{0.39}_{0.01}$ \\
R$^2$-50\% & $0.33_{0.01}$ & $\mathit{0.48}_{0.01}$ & $\underline{0.49}_{0.01}$ & $0.47_{0.01}$ & $\underline{0.49}_{0.00}$ & $\mathit{0.48}_{0.01}$ & $0.44_{0.11}$ & $0.47_{0.00}$ & $0.47_{0.01}$ & $\underline{0.49}_{0.01}$ & $\mathit{0.48}_{0.01}$ & $\mathbf{0.50}_{0.01}$ \\ \bottomrule

    \end{tabular}
    }
    \caption{FlowerFormer NAS-Bench-Graph Results. Same experimental setup as Tab.~\ref{tab:supp_ff_nb101}. }
    \label{tab:supp_ff_nbgraph}
    
\end{table*}

\begin{table*}[t!]
    \centering
    \scalebox{\scaleboxratio}{
    \begin{tabular}{lcccccccccccc} \toprule
    \textbf{NB-101} & \textbf{MLP} & \textbf{GCN} & \textbf{SAGE} & \textbf{GIN} &  \textbf{GEN} & \textbf{$k$-GNN} & \textbf{PNA} & \textbf{GATv2-H1} & \textbf{GATv2-H4} & \textbf{GT-H1} & \textbf{GT-H4} & \textbf{GATSet}\\ \midrule
KT-100 & $0.61_{0.01}$ & $\mathit{0.67}_{0.01}$ & $\mathit{0.67}_{0.03}$ & $0.64_{0.01}$ & $\mathit{0.67}_{0.01}$ & $\mathbf{0.70}_{0.01}$ & $\underline{0.69}_{0.03}$ & $\mathit{0.67}_{0.02}$ & $0.64_{0.01}$ & $\underline{0.69}_{0.02}$ & $\mathit{0.67}_{0.03}$ & $0.65_{0.02}$ \\
KT-172 & $0.66_{0.01}$ & $\mathit{0.72}_{0.00}$ & $\underline{0.73}_{0.01}$ & $0.71_{0.01}$ & $\mathit{0.72}_{0.01}$ & $0.71_{0.01}$ & $\mathit{0.72}_{0.01}$ & $\mathit{0.72}_{0.01}$ & $\underline{0.73}_{0.01}$ & $\mathbf{0.74}_{0.00}$ & $\mathbf{0.74}_{0.01}$ & $0.71_{0.01}$ \\
KT-424 & $0.70_{0.01}$ & $0.75_{0.00}$ & $\mathit{0.76}_{0.00}$ & $\mathit{0.76}_{0.01}$ & $\underline{0.77}_{0.00}$ & $\mathit{0.76}_{0.00}$ & $\mathbf{0.78}_{0.01}$ & $0.75_{0.00}$ & $0.74_{0.01}$ & $\mathit{0.76}_{0.01}$ & $\mathit{0.76}_{0.01}$ & $\mathit{0.76}_{0.01}$ \\
KT-4236 & $0.81_{0.00}$ & $\underline{0.85}_{0.00}$ & $\mathit{0.84}_{0.00}$ & $\underline{0.85}_{0.00}$ & $\underline{0.85}_{0.00}$ & $\mathbf{0.86}_{0.00}$ & $\underline{0.85}_{0.00}$ & $\mathit{0.84}_{0.00}$ & $\mathit{0.84}_{0.00}$ & $\mathit{0.84}_{0.00}$ & $\mathit{0.84}_{0.00}$ & $\underline{0.85}_{0.00}$ \\ \midrule

SRCC-100 & $0.80_{0.01}$ & $\mathit{0.86}_{0.01}$ & $\mathit{0.86}_{0.02}$ & $0.82_{0.01}$ & $\mathit{0.86}_{0.01}$ & $\mathbf{0.88}_{0.01}$ & $\mathbf{0.88}_{0.02}$ & $\mathit{0.86}_{0.01}$ & $0.83_{0.01}$ & $\underline{0.87}_{0.01}$ & $0.85_{0.03}$ & $0.84_{0.01}$ \\
SRCC-172 & $0.85_{0.01}$ & $\underline{0.90}_{0.00}$ & $\underline{0.90}_{0.00}$ & $0.88_{0.01}$ & $\mathit{0.89}_{0.01}$ & $\mathit{0.89}_{0.00}$ & $\underline{0.90}_{0.01}$ & $\underline{0.90}_{0.00}$ & $\underline{0.90}_{0.01}$ & $\mathbf{0.91}_{0.00}$ & $\mathbf{0.91}_{0.01}$ & $\mathit{0.89}_{0.01}$ \\
SRCC-424 & $0.88_{0.00}$ & $\underline{0.92}_{0.00}$ & $\mathbf{0.93}_{0.00}$ & $\mathbf{0.93}_{0.00}$ & $\mathbf{0.93}_{0.00}$ & $\mathbf{0.93}_{0.00}$ & $\mathbf{0.93}_{0.00}$ & $\underline{0.92}_{0.00}$ & $\mathit{0.91}_{0.01}$ & $\mathbf{0.93}_{0.00}$ & $\mathbf{0.93}_{0.00}$ & $\underline{0.92}_{0.00}$ \\
SRCC-4236 & $\mathit{0.95}_{0.00}$ & $\mathbf{0.97}_{0.00}$ & $\mathbf{0.97}_{0.00}$ & $\mathbf{0.97}_{0.00}$ & $\mathbf{0.97}_{0.00}$ & $\mathbf{0.97}_{0.00}$ & $\mathbf{0.97}_{0.00}$ & $\mathbf{0.97}_{0.00}$ & $\underline{0.96}_{0.00}$ & $\mathbf{0.97}_{0.00}$ & $\mathbf{0.97}_{0.00}$ & $\mathbf{0.97}_{0.00}$ \\ \midrule

MAE-100 & $0.72_{0.02}$ & $0.69_{0.02}$ & $\underline{0.65}_{0.01}$ & $\mathbf{0.64}_{0.05}$ & $\mathit{0.66}_{0.01}$ & $\mathit{0.66}_{0.02}$ & $\mathit{0.66}_{0.02}$ & $\mathit{0.66}_{0.02}$ & $0.68_{0.01}$ & $\mathit{0.66}_{0.02}$ & $0.67_{0.02}$ & $0.69_{0.02}$ \\
MAE-172 & $0.53_{0.01}$ & $0.52_{0.01}$ & $0.50_{0.01}$ & $\underline{0.48}_{0.02}$ & $\mathit{0.49}_{0.01}$ & $\mathbf{0.46}_{0.00}$ & $0.50_{0.01}$ & $0.52_{0.01}$ & $0.51_{0.01}$ & $0.50_{0.01}$ & $0.51_{0.01}$ & $0.51_{0.01}$ \\
MAE-424 & $\underline{0.46}_{0.01}$ & $0.48_{0.01}$ & $0.48_{0.01}$ & $\mathbf{0.45}_{0.00}$ & $\mathit{0.47}_{0.01}$ & $\mathit{0.47}_{0.01}$ & $\underline{0.46}_{0.01}$ & $0.48_{0.01}$ & $\underline{0.46}_{0.01}$ & $0.48_{0.01}$ & $\mathit{0.47}_{0.01}$ & $\underline{0.46}_{0.01}$ \\
MAE-4236 & $\mathbf{0.47}_{0.01}$ & $\underline{0.48}_{0.00}$ & $\mathbf{0.47}_{0.00}$ & $\underline{0.48}_{0.00}$ & $\underline{0.48}_{0.00}$ & $\mathbf{0.47}_{0.00}$ & $\mathbf{0.47}_{0.01}$ & $\mathbf{0.47}_{0.00}$ & $\mathbf{0.47}_{0.00}$ & $\underline{0.48}_{0.00}$ & $\underline{0.48}_{0.00}$ & $\mathbf{0.47}_{0.00}$ \\ \midrule

Pr@10-100 & $0.49_{0.02}$ & $\mathit{0.58}_{0.04}$ & $\underline{0.60}_{0.03}$ & $0.53_{0.02}$ & $\mathit{0.58}_{0.01}$ & $\mathbf{0.62}_{0.01}$ & $\mathit{0.58}_{0.05}$ & $0.56_{0.01}$ & $0.51_{0.03}$ & $\underline{0.60}_{0.04}$ & $0.57_{0.05}$ & $0.52_{0.03}$ \\
Pr@10-172 & $0.59_{0.01}$ & $\mathit{0.67}_{0.00}$ & $\underline{0.68}_{0.01}$ & $0.66_{0.02}$ & $\mathit{0.67}_{0.01}$ & $0.66_{0.01}$ & $0.66_{0.01}$ & $\underline{0.68}_{0.01}$ & $\underline{0.68}_{0.02}$ & $\mathbf{0.69}_{0.01}$ & $\mathbf{0.69}_{0.02}$ & $0.63_{0.02}$ \\
Pr@10-424 & $0.60_{0.01}$ & $0.69_{0.01}$ & $\mathit{0.70}_{0.01}$ & $\mathit{0.70}_{0.01}$ & $\underline{0.71}_{0.01}$ & $\underline{0.71}_{0.01}$ & $\mathbf{0.72}_{0.01}$ & $\underline{0.71}_{0.01}$ & $0.69_{0.01}$ & $\underline{0.71}_{0.01}$ & $\underline{0.71}_{0.01}$ & $0.69_{0.01}$ \\
Pr@10-4236 & $0.75_{0.01}$ & $\mathbf{0.80}_{0.00}$ & $\underline{0.79}_{0.00}$ & $\underline{0.79}_{0.00}$ & $\underline{0.79}_{0.00}$ & $\mathbf{0.80}_{0.00}$ & $\underline{0.79}_{0.00}$ & $\underline{0.79}_{0.00}$ & $\mathit{0.78}_{0.00}$ & $\underline{0.79}_{0.00}$ & $\mathit{0.78}_{0.00}$ & $\mathit{0.78}_{0.00}$ \\ %

\bottomrule

    \end{tabular}
    }
    \caption{PINAT NAS-Bench-101 Results. Rows correspond to different evaluation metrics and proportion of the dataset utilized for training. Horizontal lines demarcate different classes of metrics. For KT, SRCC and Pr, higher is better. Lower is better for MAE. %
    Results averaged across 5 random seeds.}
    \label{tab:supp_pinat_nb101}

\end{table*}

Table~\ref{tab:supp_ff_nb301} shows our FlowerFormer NB-301 results. We omit PNA from this table as it causes CUDA OOM on this dataset. Per Tab.~\ref{tab:ff_results}, this is one of the datasets where DAGFormer falls short on KT and the pattern repeats for all other metrics. SRCC and R$^2$ performance is maximized on GIN, GEN and the $k$-GNN while MAE is minimized on one-headed GATv2 and GT.

Next, Table~\ref{tab:supp_ff_nbasr} features our extended results for FlowerFormer on NB-ASR. We observe that the MLP baseline dominance on KT is continued through SRCC and to a lesser extent on MAE minimization and R$^2$ maximization. When the MLP baseline does not perform the best, performance is split across a few other GNN types, specifically GATv2. 

\begin{table*}[t!]
    \centering
    \scalebox{\scaleboxratio}{
    \begin{tabular}{lcccccccccccc} \toprule
    \textbf{NB-201} & \textbf{MLP} & \textbf{GCN} & \textbf{SAGE} & \textbf{GIN} &  \textbf{GEN} & \textbf{$k$-GNN} & \textbf{PNA} & \textbf{GATv2-H1} & \textbf{GATv2-H4} & \textbf{GT-H1} & \textbf{GT-H4} & \textbf{GATSet} \\ \midrule
KT-78 & $0.09_{0.32}$ & $\mathbf{0.29}_{0.27}$ & $\underline{0.24}_{0.19}$ & $0.03_{0.38}$ & $-0.04_{0.24}$ & $\mathit{0.21}_{0.18}$ & $0.07_{0.32}$ & $-0.18_{0.19}$ & $-0.15_{0.23}$ & $-0.04_{0.26}$ & $0.13_{0.18}$ & $-0.03_{0.28}$ \\
KT-156 & $0.51_{0.03}$ & $\mathbf{0.55}_{0.04}$ & $\underline{0.53}_{0.02}$ & $\mathbf{0.55}_{0.04}$ & $\mathbf{0.55}_{0.05}$ & $\underline{0.53}_{0.03}$ & $\underline{0.53}_{0.04}$ & $\mathbf{0.55}_{0.04}$ & $\mathit{0.52}_{0.04}$ & $\mathbf{0.55}_{0.03}$ & $\underline{0.53}_{0.03}$ & $\mathit{0.52}_{0.04}$ \\
KT-469 & $0.62_{0.01}$ & $0.66_{0.02}$ & $\mathit{0.67}_{0.01}$ & $0.66_{0.01}$ & $\mathbf{0.69}_{0.01}$ & $\underline{0.68}_{0.01}$ & $\mathit{0.67}_{0.02}$ & $\underline{0.68}_{0.01}$ & $0.64_{0.02}$ & $\mathit{0.67}_{0.02}$ & $0.66_{0.02}$ & $0.63_{0.02}$ \\
KT-781 & $0.69_{0.01}$ & $\mathit{0.72}_{0.01}$ & $\underline{0.73}_{0.01}$ & $\mathit{0.72}_{0.01}$ & $\mathbf{0.74}_{0.01}$ & $\underline{0.73}_{0.00}$ & $\underline{0.73}_{0.01}$ & $\underline{0.73}_{0.01}$ & $0.71_{0.01}$ & $\underline{0.73}_{0.01}$ & $\mathit{0.72}_{0.00}$ & $0.69_{0.02}$ \\
KT-1563 & $0.75_{0.02}$ & $\mathit{0.76}_{0.01}$ & $\underline{0.77}_{0.01}$ & $\mathit{0.76}_{0.01}$ & $\mathbf{0.79}_{0.01}$ & $\mathit{0.76}_{0.01}$ & $\mathit{0.76}_{0.01}$ & $\underline{0.77}_{0.01}$ & $0.75_{0.01}$ & $\underline{0.77}_{0.01}$ & $\mathit{0.76}_{0.01}$ & $0.74_{0.01}$ \\ \midrule

SRCC-78 & $0.12_{0.47}$ & $\mathbf{0.40}_{0.38}$ & $\underline{0.34}_{0.27}$ & $0.05_{0.54}$ & $-0.05_{0.35}$ & $\mathit{0.31}_{0.27}$ & $0.10_{0.45}$ & $-0.27_{0.27}$ & $-0.21_{0.32}$ & $-0.06_{0.37}$ & $0.19_{0.26}$ & $-0.04_{0.39}$ \\
SRCC-156 & $\mathit{0.70}_{0.04}$ & $\mathbf{0.74}_{0.05}$ & $\underline{0.72}_{0.03}$ & $\mathbf{0.74}_{0.05}$ &$\mathbf{0.74}_{0.05}$ & $\underline{0.72}_{0.03}$ & $\underline{0.72}_{0.04}$ & $\mathbf{0.74}_{0.05}$ & $\mathit{0.70}_{0.04}$ & $\mathbf{0.74}_{0.03}$ & $\underline{0.72}_{0.04}$ & $\mathit{0.70}_{0.05}$ \\
SRCC-469 & $0.80_{0.01}$ & $\mathit{0.85}_{0.01}$ & $\mathit{0.85}_{0.01}$ & $0.84_{0.01}$ & $\mathbf{0.87}_{0.01}$ & $\underline{0.86}_{0.01}$ & $\mathit{0.85}_{0.02}$ & $\underline{0.86}_{0.01}$ & $0.82_{0.02}$ & $\mathit{0.85}_{0.02}$ & $0.84_{0.02}$ & $0.81_{0.02}$ \\
SRCC-781 & $0.87_{0.00}$ & $\mathit{0.89}_{0.01}$ & $\mathit{0.89}_{0.01}$ & $\mathit{0.89}_{0.01}$ & $\mathbf{0.91}_{0.01}$ & $\underline{0.90}_{0.00}$ & $\underline{0.90}_{0.01}$ & $\underline{0.90}_{0.00}$ & $0.88_{0.01}$ & $\mathit{0.89}_{0.01}$ & $\mathit{0.89}_{0.00}$ & $0.87_{0.02}$ \\
SRCC-1563 & $\mathit{0.91}_{0.01}$ & $\underline{0.92}_{0.00}$ & $\underline{0.92}_{0.00}$ & $\underline{0.92}_{0.00}$ & $\mathbf{0.93}_{0.01}$ & $\underline{0.92}_{0.00}$ & $\underline{0.92}_{0.01}$ & $\underline{0.92}_{0.00}$ & $\mathit{0.91}_{0.01}$ & $\underline{0.92}_{0.00}$ & $\underline{0.92}_{0.01}$ & $\mathit{0.91}_{0.01}$ \\ \midrule

MAE-78 & $0.60_{0.14}$ & $\mathbf{0.49}_{0.09}$ & $0.56_{0.10}$ & $0.59_{0.17}$ & $0.59_{0.10}$ & $0.55_{0.09}$ & $0.60_{0.18}$ & $\underline{0.50}_{0.04}$ & $\mathit{0.51}_{0.09}$ & $\underline{0.50}_{0.06}$ & $0.63_{0.21}$ & $0.58_{0.15}$ \\
MAE-156 & $0.36_{0.03}$ & $\underline{0.32}_{0.02}$ & $\underline{0.32}_{0.03}$ & $0.34_{0.04}$ & $\mathit{0.33}_{0.03}$ & $\underline{0.32}_{0.03}$ & $0.34_{0.02}$ & $0.34_{0.02}$ & $0.36_{0.03}$ & $\mathbf{0.31}_{0.01}$& $\mathit{0.33}_{0.02}$ & $0.34_{0.02}$ \\
MAE-469 & $0.23_{0.01}$ & $\mathit{0.20}_{0.03}$ & $\mathbf{0.18}_{0.01}$ & $\mathit{0.20}_{0.01}$ & $\underline{0.19}_{0.01}$ & $\underline{0.19}_{0.01}$ & $\underline{0.19}_{0.01}$ & $\underline{0.19}_{0.01}$ & $0.21_{0.01}$ & $\mathbf{0.18}_{0.01}$ & $\underline{0.19}_{0.01}$ & $0.22_{0.02}$ \\
MAE-781 & $0.18_{0.01}$ & $\mathit{0.15}_{0.02}$ & $\mathbf{0.13}_{0.00}$ & $\underline{0.14}_{0.01}$ & $\mathbf{0.13}_{0.01}$ & $\mathbf{0.13}_{0.01}$ & $\mathbf{0.13}_{0.01}$ & $\underline{0.14}_{0.01}$ & $0.16_{0.01}$ & $\mathbf{0.13}_{0.00}$ & $\underline{0.14}_{0.01}$ & $0.17_{0.01}$ \\
MAE-1563 & $\underline{0.12}_{0.01}$ & $\mathbf{0.10}_{0.00}$ & $\mathbf{0.10}_{0.00}$ & $\mathbf{0.10}_{0.00}$ & $\mathbf{0.10}_{0.00}$ & $\mathbf{0.10}_{0.00}$ & $\mathbf{0.10}_{0.00}$ & $\mathbf{0.10}_{0.00}$ & $\underline{0.12}_{0.01}$ & $\mathbf{0.10}_{0.00}$ & $\mathbf{0.10}_{0.00}$ & $\underline{0.12}_{0.01}$ \\ \midrule

Pr@10-78 & $0.18_{0.19}$ & $\mathbf{0.33}_{0.17}$ & $\underline{0.25}_{0.15}$ & $0.18_{0.15}$ & $0.10_{0.10}$ & $\mathit{0.22}_{0.15}$ & $0.18_{0.19}$ & $0.02_{0.03}$ & $0.05_{0.10}$ & $0.07_{0.10}$ & $0.17_{0.11}$ & $0.10_{0.17}$ \\
Pr@10-156 & $0.38_{0.02}$ & $\underline{0.41}_{0.10}$ & $0.37_{0.06}$ & $\underline{0.41}_{0.06}$ & $\mathbf{0.46}_{0.07}$ & $0.36_{0.08}$ & $0.37_{0.09}$ & $0.39_{0.11}$ & $0.35_{0.05}$ & $\mathit{0.40}_{0.06}$ & $0.39_{0.07}$ & $0.35_{0.05}$ \\
Pr@10-469 & $0.44_{0.05}$ & $0.51_{0.02}$ & $0.50_{0.04}$ & $\mathit{0.52}_{0.05}$ & $\mathbf{0.58}_{0.04}$ & $0.51_{0.03}$ & $0.51_{0.06}$ & $\underline{0.54}_{0.04}$ & $0.48_{0.05}$ & $\mathit{0.52}_{0.02}$ & $0.51_{0.05}$ & $0.48_{0.05}$ \\
Pr@10-781 & $0.55_{0.02}$ & $\mathit{0.59}_{0.02}$ & $0.58_{0.04}$ & $\mathit{0.59}_{0.04}$ & $\mathbf{0.63}_{0.03}$ & $0.58_{0.03}$ & $\underline{0.60}_{0.05}$ & $0.58_{0.03}$ & $0.58_{0.03}$ & $0.58_{0.03}$ & $\mathit{0.59}_{0.01}$ & $\mathit{0.59}_{0.04}$ \\
Pr@10-1563 & $\mathit{0.63}_{0.03}$ & $\mathit{0.63}_{0.03}$ & $\underline{0.64}_{0.01}$ & $\underline{0.64}_{0.01}$ & $\mathbf{0.67}_{0.00}$ & $\underline{0.64}_{0.01}$ & $\underline{0.64}_{0.02}$ & $0.62_{0.02}$ & $0.62_{0.03}$ & $\mathit{0.63}_{0.02}$ & $\mathit{0.63}_{0.01}$ & $\mathit{0.63}_{0.02}$ \\ %

\bottomrule

    \end{tabular}
    }
    \caption{PINAT NAS-Bench-201 Results. Same setup as Tab.~\ref{tab:supp_pinat_nb101}.}
    \label{tab:supp_pinat_nb201}
    
\end{table*}

\begin{table*}[t!]
    \centering
    \scalebox{\scaleboxratio}{
    \begin{tabular}{lcccccccccccc} \toprule
    \textbf{NB-101} & \textbf{MLP} & \textbf{GCN} & \textbf{SAGE} & \textbf{GIN} &  \textbf{GEN} & \textbf{$k$-GNN} & \textbf{PNA} & \textbf{GATv2-H1} & \textbf{GATv2-H4} & \textbf{GT-H1} & \textbf{GT-H4} \\ \midrule
MAE & $0.34_{0.01}$ & $0.22_{0.02}$ & $\underline{0.16}_{0.00}$ & $\mathit{0.18}_{0.01}$ & $\mathbf{0.15}_{0.00}$ & $\mathbf{0.15}_{0.00}$ & $\mathbf{0.15}_{0.01}$ & $0.20_{0.00}$ & $\underline{0.16}_{0.01}$ & $\underline{0.16}_{0.01}$ & $\mathbf{0.15}_{0.01}$ \\
MAPE & $4.80_{0.94}$ & $2.81_{0.36}$ & $1.97_{0.13}$ & $2.24_{0.42}$ & $\underline{1.75}_{0.32}$ & $\mathbf{1.69}_{0.14}$ & $\underline{1.75}_{0.30}$ &  $2.48_{0.24}$ & $1.95_{0.19}$ & $2.03_{0.32}$ & $\mathit{1.80}_{0.29}$ \\
SRCC & $0.68_{0.00}$ & $0.88_{0.02}$ & $\underline{0.96}_{0.00}$ & $0.93_{0.01}$ & $\underline{0.96}_{0.00}$ & $\underline{0.96}_{0.00}$ & $\mathbf{0.97}_{0.00}$ &  $0.91_{0.01}$ & $\mathit{0.95}_{0.01}$ & $\underline{0.96}_{0.01}$ & $\underline{0.96}_{0.00}$ \\
KT & $0.49_{0.00}$ & $0.70_{0.03}$ & $\mathit{0.82}_{0.01}$ & $0.78_{0.01}$ & $\underline{0.83}_{0.01}$ & $\underline{0.83}_{0.00}$ & $\mathbf{0.84}_{0.00}$ &  $0.74_{0.01}$ & $0.81_{0.01}$ & $\mathit{0.82}_{0.01}$ & $\underline{0.83}_{0.01}$ \\ \bottomrule
    \end{tabular}
    }
    \caption{AutoBuild NAS-Bench-101 results predicting $y_\mathcal{G}$. Rows correspond to different evaluation metrics. For MAE and MAPE, lower is better. For SRCC and KT higher is better. 
    Results averaged across 5 random seeds.}
    \label{tab:supp_autobuild_nb101}

\end{table*}

\begin{table*}[t!]
    \centering
    \scalebox{\scaleboxratio}{
    \begin{tabular}{lcccccccccccc} \toprule
    \textbf{NB-201} & \textbf{MLP} & \textbf{GCN} & \textbf{SAGE} & \textbf{GIN} &  \textbf{GEN} & \textbf{$k$-GNN} & \textbf{PNA} & \textbf{GATv2-H1} & \textbf{GATv2-H4} & \textbf{GT-H1} & \textbf{GT-H4} \\ \midrule
MAE & $0.21_{0.03}$ & $0.40_{0.25}$ & $0.15_{0.05}$ & $\mathit{0.13}_{0.02}$ & $\mathbf{0.07}_{0.01}$ & $0.15_{0.05}$ & $\underline{0.09}_{0.03}$ & $0.21_{0.08}$ & $0.49_{0.23}$ & $\underline{0.09}_{0.02}$ & $\underline{0.09}_{0.02}$ \\
MAPE & $2.66_{1.33}$ & $6.51_{4.23}$ & $4.97_{6.43}$ & $2.33_{1.04}$ & $\mathbf{1.31}_{0.68}$ & $7.84_{12.41}$ & $\underline{1.75}_{1.08}$ & $7.68_{10.25}$ & $11.39_{11.49}$ & $2.98_{3.53}$ & $\mathit{1.91}_{1.74}$ \\
SRCC & $0.85_{0.01}$ & $0.76_{0.08}$ & $0.88_{0.04}$ & $0.86_{0.02}$ & $\mathbf{0.95}_{0.01}$ & $0.89_{0.02}$ & $\underline{0.93}_{0.03}$ &  $0.80_{0.07}$ & $0.73_{0.08}$ & $\mathit{0.92}_{0.02}$ & $\underline{0.93}_{0.02}$ \\
KT & $0.67_{0.01}$ & $0.57_{0.08}$ & $0.70_{0.04}$ & $0.69_{0.03}$ & $\mathbf{0.81}_{0.02}$ & $0.72_{0.03}$ & $\underline{0.79}_{0.04}$ & $0.62_{0.08}$ & $0.54_{0.07}$ & $0.76_{0.03}$ & $\mathit{0.78}_{0.04}$ \\ \bottomrule
    \end{tabular}
    }
    \caption{AutoBuild NAS-Bench-201 results predicting $y_\mathcal{G}$. Same setup as Tab.~\ref{tab:supp_autobuild_nb101}.}
    \label{tab:supp_autobuild_nb201}

\end{table*}

\begin{table*}[t!]
    \centering
    \scalebox{\scaleboxratio}{
    \begin{tabular}{lcccccccccccc} \toprule
    \textbf{NB-301} & \textbf{MLP} & \textbf{GCN} & \textbf{SAGE} & \textbf{GIN} &  \textbf{GEN} & \textbf{$k$-GNN} & \textbf{PNA} & \textbf{GATv2-H1} & \textbf{GATv2-H4} & \textbf{GT-H1} & \textbf{GT-H4} \\ \midrule
MAE & $\mathit{0.27}_{0.01}$ & $0.43_{0.09}$ & $0.29_{0.05}$ & $\mathit{0.27}_{0.04}$ & $\mathbf{0.25}_{0.01}$ & $\underline{0.26}_{0.02}$ & $0.32_{0.09}$ & $0.33_{0.04}$ & $0.34_{0.10}$ & $0.28_{0.01}$ & $\mathit{0.27}_{0.02}$ \\
MAPE & $2.75_{1.08}$ & $4.98_{1.80}$ & $3.15_{1.96}$ & $\underline{2.25}_{0.53}$ & $\mathbf{2.18}_{0.64}$ & $2.53_{0.94}$ & $5.35_{6.35}$ & $4.07_{3.65}$ & $2.92_{1.19}$ & $\mathit{2.35}_{0.36}$ & $3.38_{1.94}$ \\
SRCC & $0.85_{0.00}$ & $0.85_{0.02}$ & $\mathit{0.86}_{0.01}$ & $\underline{0.87}_{0.00}$ & $\mathbf{0.88}_{0.00}$ & $\mathit{0.86}_{0.01}$ & $\underline{0.87}_{0.01}$ &  $0.85_{0.01}$ & $\underline{0.87}_{0.01}$ & $\mathit{0.86}_{0.01}$ & $\underline{0.87}_{0.01}$ \\
KT & $0.67_{0.01}$ & $0.67_{0.03}$ & $\mathit{0.68}_{0.02}$ & $\mathbf{0.70}_{0.00}$ & $\mathbf{0.70}_{0.01}$ & $\underline{0.69}_{0.01}$ & $\underline{0.69}_{0.01}$ & $0.66_{0.01}$ & $\underline{0.69}_{0.01}$ & $\mathit{0.68}_{0.01}$ & $\underline{0.69}_{0.01}$ \\ \bottomrule
    \end{tabular}
    }
    \caption{AutoBuild NAS-Bench-301 results predicting $y_\mathcal{G}$. Same setup as Tab.~\ref{tab:supp_autobuild_nb101}.}
    \label{tab:supp_autobuild_nb301}

\end{table*}

Finally, Table~\ref{tab:supp_ff_nbgraph} enumerates FlowerFormer %
on NB-Graph. The dominance DAGFormer held on this benchmark for KT is reflected in all other metrics as the only time it does not achieve the best performance is on KT-50\%, where best performance is shared amongst other, more hardware-friendly GNNs, and SRCC-50\%, where GCN and GT-H4 achieve the best ranking correlation, while GraphSAGE, GEN, GATv2-H1, GATv2-H4 and GT-H1 tie for 2nd place.

\subsubsection{PINAT.}
Table~\ref{tab:supp_pinat_nb101} shows our results for PINAT on NAS-Bench-101 spanning additional metrics like SRCC, MAE, and Precision (Pr) at several levels, in addition to KT. Considering SRCC as a ranking metric over KT opens up the opportunity for many ties in performance for first place, especially as the size of the training data increases. Now GCN, GraphSAGE, GIN, GEN, GATv2, GT and potentially GATSet tie with $k$-GNN and PNA depending on the data split. However, for MAE performance only GIN and $k$-GNN achieve the best performance more than once, while there are multiple first place ties when the size of the training data is maximized. Further, we also consider Precision @ $K=10$ which focuses not on global dataset performance but, like NDCG, the ability of the predictor to tell if a data sample has a high label. Here, we see mixed results where $k$-GNN achieves the best Pr@10 when the data split is the largest and smallest, but, like KT, the middle partitions are won by PNA and GT. The GCN also ties the $k$-GNN at the largest split.

Finally, Table~\ref{tab:supp_pinat_nb201} displays our PINAT results for NAS-Bench-201. The SRCC performance largely mirrors that of KT: If a GNN type was the best for a given trial before for KT, it also achieves the best SRCC, though there are several more ties for 2nd and 3rd place performance. Precision @ $K=10$ performance is largest similar with a few minor adjustments to places and the complete loss of GT performance to GCN and GEN. 

\begin{table*}[t!]
    \centering
    \scalebox{\scaleboxratio}{
    \begin{tabular}{lcccccccccccc} \toprule
    \textbf{OFA-MBv3} & \textbf{MLP} & \textbf{GCN} & \textbf{SAGE} & \textbf{GIN} &  \textbf{GEN} & \textbf{$k$-GNN} & \textbf{PNA} & \textbf{GATv2-H1} & \textbf{GATv2-H4} & \textbf{GT-H1} & \textbf{GT-H4} \\ \midrule
MAE & $\underline{0.30}_{0.04}$ & $0.34_{0.03}$ & $0.32_{0.05}$ & $0.39_{0.08}$ & $0.39_{0.08}$ & $0.32_{0.05}$ & $0.33_{0.05}$ & $0.35_{0.04}$ & $0.32_{0.05}$ & $\mathit{0.31}_{0.04}$ & $\mathbf{0.29}_{0.05}$ \\
MAPE & $\mathit{20.31}_{41.61}$ & $\underline{18.89}_{35.31}$ & $41.12_{79.95}$ & $53.80_{105.00}$ & $\mathbf{18.41}_{33.67}$ & $41.17_{80.05}$ & $79.31_{155.55}$ &  $29.25_{56.17}$ & $32.90_{63.51}$ & $43.14_{83.79}$ & $22.37_{42.61}$ \\
SRCC & $\underline{0.92}_{0.02}$ & $\mathit{0.91}_{0.02}$ & $\mathit{0.91}_{0.03}$ & $0.86_{0.06}$ & $0.87_{0.05}$ & $\mathit{0.91}_{0.03}$ & $\mathit{0.91}_{0.03}$ &  $0.90_{0.03}$ & $\mathit{0.91}_{0.03}$ & $\underline{0.92}_{0.02}$ & $\mathbf{0.93}_{0.03}$ \\
KT & $\mathbf{0.77}_{0.04}$ & $0.74_{0.03}$ & $\mathit{0.75}_{0.06}$ & $0.69_{0.08}$ & $0.69_{0.07}$ & $\mathit{0.75}_{0.05}$ & $\mathit{0.75}_{0.04}$ &  $0.73_{0.04}$ & $\mathit{0.75}_{0.05}$ & $\underline{0.76}_{0.04}$ & $\mathbf{0.77}_{0.04}$ \\ \bottomrule
    \end{tabular}
    }
    \caption{AutoBuild OFA-MBv3 results predicting $y_\mathcal{G}$. Same setup as Tab.~\ref{tab:supp_autobuild_nb101}.}
    \label{tab:supp_autobuild_mbv3}

\end{table*}

\begin{table*}[t!]
    \centering
    \scalebox{\scaleboxratio}{
    \begin{tabular}{lcccccccccccc} \toprule
    \textbf{OFA-PN} & \textbf{MLP} & \textbf{GCN} & \textbf{SAGE} & \textbf{GIN} &  \textbf{GEN} & \textbf{$k$-GNN} & \textbf{PNA} & \textbf{GATv2-H1} & \textbf{GATv2-H4} & \textbf{GT-H1} & \textbf{GT-H4} \\ \midrule
MAE & $\mathit{0.15}_{0.02}$ & $0.19_{0.01}$ & $\mathbf{0.12}_{0.02}$ & $0.19_{0.02}$ & $\underline{0.14}_{0.01}$ & $\mathbf{0.12}_{0.02}$ & $0.16_{0.05}$ & $0.17_{0.02}$ & $0.16_{0.02}$ & $\underline{0.14}_{0.02}$ & $\underline{0.14}_{0.02}$ \\
MAPE & $\underline{0.76}_{0.15}$ & $1.24_{0.37}$ & $0.83_{0.17}$ & $\mathbf{0.62}_{0.27}$ & $2.20_{3.32}$ & $1.32_{1.36}$ & $0.81_{0.41}$ & $0.80_{0.20}$ & $2.60_{2.56}$ & $0.84_{0.22}$ & $\mathit{0.78}_{0.39}$ \\
SRCC & $\underline{0.98}_{0.00}$ & $\underline{0.98}_{0.00}$ & $\underline{0.98}_{0.00}$ & $\mathbf{0.99}_{0.00}$ & $\mathbf{0.99}_{0.00}$ & $\mathbf{0.99}_{0.00}$ & $\mathbf{0.99}_{0.00}$ & $\underline{0.98}_{0.00}$ & $\underline{0.98}_{0.01}$ & $\mathbf{0.99}_{0.00}$ & $\mathbf{0.99}_{0.00}$ \\
KT & $0.88_{0.01}$ & $0.87_{0.00}$ & $\mathit{0.90}_{0.01}$ & $\mathbf{0.92}_{0.01}$ & $\mathbf{0.92}_{0.01}$ & $\mathit{0.90}_{0.01}$ & $\underline{0.91}_{0.01}$ & $0.89_{0.01}$ & $0.88_{0.03}$ & $\underline{0.91}_{0.00}$ & $\mathbf{0.92}_{0.00}$ \\ \bottomrule
    \end{tabular}
    }
    \caption{AutoBuild OFA-PN results predicting $y_\mathcal{G}$. Same setup as Tab.~\ref{tab:supp_autobuild_nb101}.}
    \label{tab:supp_autobuild_pn}

\end{table*}

\begin{table*}[t!]
    \centering
    \scalebox{\scaleboxratio}{
    \begin{tabular}{lcccccccccccc} \toprule
    \textbf{ONNX-IR} & \textbf{MLP} & \textbf{GCN} & \textbf{SAGE} & \textbf{GIN} &  \textbf{GEN} & \textbf{$k$-GNN} & \textbf{PNA} & \textbf{GATv2-H1} & \textbf{GATv2-H4} & \textbf{GT-H1} & \textbf{GT-H4} \\ \midrule
MAE & $0.19_{0.04}$ & $0.27_{0.09}$ & $0.18_{0.06}$ & $\mathbf{0.12}_{0.02}$ & $\mathit{0.15}_{0.03}$ & $\mathit{0.15}_{0.03}$ & $\underline{0.13}_{0.03}$ & $0.18_{0.03}$ & $0.57_{0.62}$ & $\mathbf{0.12}_{0.01}$ & $\underline{0.13}_{0.03}$ \\
MAPE & $4.87_{4.23}$ & $1.16_{0.55}$ & $1.45_{0.83}$ & $\mathit{1.14}_{0.23}$ & $\mathbf{0.67}_{0.21}$ & $\underline{0.68}_{0.16}$ & $1.20_{0.49}$ & $1.18_{0.62}$ & $1.36_{0.50}$ & $2.35_{3.11}$ & $2.15_{2.65}$ \\
SRCC & $\underline{0.98}_{0.00}$ & $\mathit{0.97}_{0.01}$ & $\mathbf{0.99}_{0.01}$ & $\mathit{0.97}_{0.01}$ & $\mathbf{0.99}_{0.00}$ & $\mathbf{0.99}_{0.00}$ & $\mathbf{0.99}_{0.00}$ & $\underline{0.98}_{0.01}$ & $\mathbf{0.99}_{0.00}$ & $\underline{0.98}_{0.01}$ & $\mathbf{0.99}_{0.00}$ \\
KT & $\mathit{0.89}_{0.01}$ & $0.85_{0.02}$ & $\underline{0.90}_{0.02}$ & $0.86_{0.02}$ & $\underline{0.90}_{0.01}$ & $\underline{0.90}_{0.01}$ & $\mathbf{0.91}_{0.01}$ & $0.87_{0.02}$ & $\underline{0.90}_{0.01}$ & $\mathit{0.89}_{0.02}$ & $\underline{0.90}_{0.01}$ \\ \bottomrule
    \end{tabular}
    }
    \caption{AutoBuild MBV3-PN-ONNX-IR results predicting $y_\mathcal{G}$. Same setup as Tab.~\ref{tab:supp_autobuild_nb101}.}
    \label{tab:supp_autobuild_onnx}

\end{table*}

\begin{table*}[t!]
    \centering
    \scalebox{\scaleboxratio}{
    \begin{tabular}{lcccccccccccc} \toprule
    \textbf{PixArt-$\alpha$} & \textbf{MLP} & \textbf{GCN} & \textbf{SAGE} & \textbf{GIN} &  \textbf{GEN} & \textbf{$k$-GNN} & \textbf{PNA} & \textbf{GATv2-H1} & \textbf{GATv2-H4} & \textbf{GT-H1} & \textbf{GT-H4} \\ \midrule
    SRCC & $\mathit{0.29}_{0.03}$ & $0.02_{0.02}$ & $0.12_{0.04}$ & $-0.01_{0.03}$ & $\mathbf{0.36}_{0.03}$ & $0.12_{0.03}$ & $\underline{0.35}_{0.04}$ & $-0.30_{0.06}$ & $0.15_{0.02}$ & $\mathit{0.29}_{0.01}$ & $-0.13_{0.04}$ \\
    NDCG & $\mathit{0.58}_{0.05}$ & $0.41_{0.03}$ & $0.43_{0.03}$ & $0.19_{0.11}$ & $\underline{0.61}_{0.05}$ & $0.52_{0.05}$ & $\underline{0.61}_{0.03}$ & $0.18_{0.13}$ & $0.50_{0.03}$ & $\mathbf{0.72}_{0.09}$ & $0.25_{0.05}$ \\

    KT & $\underline{0.21}_{0.02}$ & $0.01_{0.02}$ & $0.08_{0.03}$ & $-0.01_{0.02}$ & $\mathbf{0.25}_{0.02}$ & $0.09_{0.02}$ & $\mathbf{0.25}_{0.03}$ & $-0.21_{0.04}$ & $\mathit{0.11}_{0.02}$ & $\underline{0.21}_{0.01}$ & $-0.10_{0.03}$ \\ 
    MAE & $17.13_{0.65}$ & $17.18_{0.55}$ & $17.14_{0.57}$ & $17.13_{0.54}$ &$17.15_{0.38}$ & $\underline{17.08}_{0.37}$ & $17.15_{0.51}$ & $17.13_{0.63}$ & $\mathit{17.11}_{0.33}$ & $\mathbf{17.07}_{0.67}$ & $17.12_{0.80}$ \\ \bottomrule
    \end{tabular}
    }
    \caption{Qua$^2$SeDiMo PixArt-$\alpha$ results predicting $y_\mathcal{G}$. Rows demarcate different metrics. For SRCC, NDCG and KT, higher is better. For MAE, lower is better. Results averaged across 5 seeds. }
    \label{tab:supp_qua2sedimo_alpha}
    \vspace{-3mm}
\end{table*}

\begin{table*}[t!]
    \centering
    \scalebox{\scaleboxratio}{
    \begin{tabular}{lccccccccccccccc} \toprule
    \textbf{Hunyuan} & \textbf{MLP} & \textbf{GCN} & \textbf{SAGE} & \textbf{GIN} &  \textbf{GEN} & \textbf{$k$-GNN} & \textbf{PNA} & \textbf{GATv2-H1} & \textbf{GATv2-H4} & \textbf{GT-H1} & \textbf{GT-H4} \\ \midrule
    SRCC & $\underline{0.42}_{0.03}$ & $0.10_{0.03}$ & $-0.13_{0.04}$ & $-0.06_{0.02}$ & $\mathit{0.38}_{0.03}$ & $-0.17_{0.01}$ & $\mathbf{0.73}_{0.02}$ & $0.10_{0.07}$ & $-0.48_{0.03}$ & $-0.19_{0.01}$ & $-0.48_{0.04}$ \\
    NDCG & $\underline{0.64}_{0.02}$ & $0.46_{0.07}$ & $0.36_{0.07}$ & $0.40_{0.03}$& $\mathit{0.60}_{0.04}$ & $0.39_{0.04}$ & $\mathbf{0.85}_{0.07}$ & $0.39_{0.04}$ & $0.18_{0.09}$ & $0.24_{0.10}$ & $0.19_{0.05}$ \\

    KT & $\underline{0.31}_{0.02}$ & $0.07_{0.02}$ & $-0.09_{0.03}$ & $-0.03_{0.02}$ & $\mathit{0.28}_{0.03}$ & $-0.12_{0.01}$ & $\mathbf{0.55}_{0.02}$ & $0.08_{0.05}$ & $-0.35_{0.02}$ & $-0.14_{0.01}$ & $-0.37_{0.04}$ \\ 
    MAE & $19.19_{0.19}$ & $19.27_{0.81}$ & $\mathbf{19.10}_{0.26}$ & $19.19_{0.53}$ & $19.22_{0.72}$ & $\mathbf{19.10}_{0.52}$ & $\underline{19.12}_{0.40}$ & $\mathit{19.15}_{0.80}$ & $19.20_{0.27}$ & $19.17_{0.60}$ & $19.24_{0.24}$ \\\bottomrule
    \end{tabular}
    }
    \caption{Qua$^2$SeDiMo Hunyuan results predicting $y_\mathcal{G}$. Rows demarcate different metrics. Results averaged across 5 seeds. }
    \label{tab:supp_qua2sedimo_hunyuan}
\end{table*}

\begin{table*}[t!]
    \centering
    \scalebox{\scaleboxratio}{
    \begin{tabular}{lcccccccccccccc} \toprule
    \textbf{SDXL} & \textbf{GCN} & \textbf{SAGE} & \textbf{GIN} &  \textbf{GEN} & \textbf{$k$-GNN} & \textbf{PNA} & \textbf{GATv2-H1} & \textbf{GATv2-H4} & \textbf{GT-H1} & \textbf{GT-H4} \\ \midrule
    SRCC & $-0.08_{0.04}$ & $0.01_{0.04}$ & $0.09_{0.03}$ & $\mathit{0.42}_{0.02}$ & $-0.09_{0.03}$ & $\underline{0.67}_{0.03}$ & $0.41_{0.02}$ & $\mathbf{0.68}_{0.03}$ & $-0.09_{0.01}$ & $0.26_{0.03}$ \\
    NDCG & $0.43_{0.04}$ & $0.43_{0.02}$ & $0.40_{0.03}$ & $0.63_{0.04}$ & $0.34_{0.03}$ & $\underline{0.86}_{0.04}$ & $\mathit{0.66}_{0.01}$ & $\mathbf{0.89}_{0.04}$ & $0.36_{0.06}$ & $0.51_{0.01}$ \\
    KT & $-0.05_{0.04}$ & $0.01_{0.03}$ & $0.07_{0.02}$ & $\mathit{0.31}_{0.01}$ & $-0.06_{0.03}$ & $\underline{0.49}_{0.03}$ & $0.30_{0.02}$ & $\mathbf{0.50}_{0.03}$ & $-0.07_{0.00}$ & $0.19_{0.02}$ \\
    MAE & $\underline{17.82}_{0.46}$ & $17.96_{0.56}$ & $17.91_{0.73}$ & $17.92_{0.60}$ & $17.88_{0.20}$ & $17.91_{0.24}$ & $\mathit{17.84}_{0.19}$ & $\mathbf{7.96}_{0.16}$ & $\mathit{17.84}_{0.45}$ & $17.91_{0.40}$ \\ \bottomrule
    \end{tabular}
    }
    \caption{Qua$^2$SeDiMo SDXL results predicting $y_\mathcal{G}$. Rows demarcate different metrics. Results averaged across 5 seeds. Note that the MLP does not work on this dataset and produces NaNs.}
    \label{tab:supp_qua2sedimo_sdxl}
    \vspace{-4mm}
\end{table*}

\subsubsection{AutoBuild End-to-End Prediction.}
FlowerFormer and PINAT are both end-to-end predictors, i.e., their sole goal is accurate estimation of $y_\mathcal{G}$. AutoBuild facilitates this as well but also performs hop-level prediction. This subsection covers prediction of $y_\mathcal{G}$.

Table~\ref{tab:supp_autobuild_nb101} covers NAS-Bench-101. The GEN, $k$-GNN and PNA achieve high performance across all metrics, while GT-H4 ties for best MAE performance. GraphSAGE and GT-H1 also do well on several metrics.

Next, Table~\ref{tab:supp_autobuild_nb201} covers NAS-Bench-201 which is dominated by GEN for best performance on all metrics, error regression and rank correlation alike. PNA mostly dominates 2nd place but is tied by GT-H1 and GT-H4 on MAE and GT-H4 on SRCC.

Table~\ref{tab:supp_autobuild_nb301} covers NAS-Bench-301 with a similar story of GEN dominance for first place along all metrics, though sharing a tie with GIN for Kendall's Tau performance. 2nd place is scattered between the GIN, $k$-GNN, PNA and -H4 attention-based GNNs.

Next, Table~\ref{tab:supp_autobuild_mbv3} covers OFA-MBv3, the first of two sequence graph datasets. Surprisingly, GT-H4 sweeps each metric except for Mean Absolute Percentage Error (MAPE) which is minimized by GEN. Also, due to the sequence graph nature of this dataset with predictable topology, the MLP baseline ties GT-H4 for KT performance and does well on MAE and SRCC as well.

Also, we have Table~\ref{tab:supp_autobuild_pn} which covers OFA-PN. GT-H4 achieves the best ranking metrics SRCC and KT but this performance is tied with GIN and GEN. GIN specifically achieves the best result on all ranking metrics, whereas the only other GNNs to get first place performance more than once are GEN, $k$-GNN and GT-H1.

Finally, Table~\ref{tab:supp_autobuild_onnx} covers the ONNX-IR representation of MBv3 and PN; these graphs describe the same DNNs but are much larger and complex. Here, performance is scattered. GIN and GT-H1 minimize MAE performance, 
whereas GEN achieves the best MAPE, and $k$-GNN achieves the 2nd best. Several GNN types tie for near perfect SRCC performance and several achieve 0.9 or better KT, with PNA edging out all other GNN types slightly.

\subsubsection{Qua$^2$SeDiMo End-to-End Prediction.}

Table~\ref{tab:supp_qua2sedimo_alpha} covers the end-to-end prediction performance for Qua$^2$SeDiMo on PixArt-$\alpha$ across four metrics. GEN achieves the best SRCC and KT ranking performance with PNA tied on KT and a close second for SRCC. Both of these fall 2nd place to GT-H1 for NDCG performance which is more concerned with identifying the graphs with the highest labels $y_\mathcal{G}$ as opposed to global ranking performance, as well as MAE regression performance.

Next, Table~\ref{tab:supp_qua2sedimo_hunyuan} repeats these findings for Hunyuan. PNA achieves the best SRCC, NDCG and KT metrics while the $k$-GNN is tied with GraphSAGE for the best MAE. It is also worth noting just how wildly the SRCC and KT values for some GNNs can swing on this dataset, depending on the choice of GNN layer, from strong positive correlation (0.73 SRCC/0.55 KT) to very negative (-0.48 SRCC/-0.35 KT).

Finally, Table~\ref{tab:supp_qua2sedimo_sdxl} covers SDXL. GATv2-H4 dominates across all metrics - SRCC, NDCG, KT and MAE. Specifically, while the results on SRCC, NDCG and KT are close with PNA taking a near 2nd place, the MAE performance is sharp as GATv2-H4 reports a markedly lower MAE (7.96); we note this as an outlier relative to all other layers (17.8-18.0). %

\subsection{Hardware and Software Utilized}
All experiments were performed on an Ubuntu 24.04 LTS workstation computer with an Intel Xeon w7-2495X, 2x Nvidia RTX Pro 5000 48GB GPUs and 512GB of RAM. GNN predictors are implemented in %
PyTorch and PyTorch-Geometric~\cite{Fey/Lenssen/2019}.

\end{document}